\pdfoutput=1
\documentclass[11pt]{article}
\usepackage{amsmath}
\usepackage{pifont}
\usepackage{amsfonts} 
\usepackage{times}
\usepackage{latexsym}
\usepackage{microtype}
\usepackage{inconsolata}
\usepackage{kotex}
\usepackage[utf8]{inputenc}
\usepackage[T1]{fontenc}
\usepackage{graphicx}
\usepackage{tabularx}
\usepackage{caption}
\usepackage{booktabs}
\usepackage{multicol}
\usepackage{multirow}
\usepackage{xcolor}
\usepackage{adjustbox}    
\usepackage{float}
\usepackage{subcaption}
\newcommand{\mj}[1]{\textcolor{black}{#1}}
\usepackage{authblk} 
\usepackage[review]{acl}

\title{X-LLaVA: Optimizing Bilingual Large Vision-Language Alignment} 

\author{
     \textbf{Dongjae Shin$^{\ddagger}$\thanks{*These authors contributed equally.} , Hyeonseok Lim$^{*}$, Inho Won$^{\ddagger}$, Changsu Choi, Minjun Kim,}\\
    \textbf{Seungwoo Song, Hangyeol Yoo, Sangmin Kim, Kyungtae Lim\thanks{$^{\dagger}$Corresponding author.}} \\ 
    Seoul National University of Science and Technology \\
    $^{\ddagger}$Teddysum \\
    \fontfamily{qcr}\selectfont
    \{dylan1998,gustjrantk,wih1226,choics2623,mjkmain\}@seoultech.ac.kr \\
    \fontfamily{qcr}\selectfont
    \{sswoo,21102372,sangmin6600,ktlim\}@seoultech.ac.kr
    } 
\begin{document}
\nolinenumbers
{\makeatletter\acl@finalcopytrue
  \maketitle
}

\begin{abstract}

The impressive development of large language models (LLMs) is expanding into the realm of large multimodal models (LMMs), \mj{which incorporate multiple types of data beyond text.}  However, the nature of multimodal models leads to significant expenses in the creation of training data. \mj{Furthermore, constructing multilingual data for LMMs presents its own set of challenges due to language diversity and complexity.} Therefore, in this study, we propose two cost-effective methods to solve this problem: (1) vocabulary expansion and pretraining of multilingual LLM for specific languages, and (2) automatic \mj{and elaborate} construction of multimodal datasets using GPT4-V. Based on these methods, we constructed a 91K English-Korean-Chinese multilingual, multimodal training dataset. Additionally, we developed a bilingual multimodal model that exhibits excellent performance in both Korean and English, \mj{surpassing existing approaches.}

\end{abstract}

\section{Introduction} \label{sec:introduction}
Recently, large multimodal models (LMMs) have evolved to respond in alignment with human intent through visual instruction-following (VIF) \cite{liu2023improved, dai2023instructblip, Qwen-VL, chen2023visual, openai2023gpt4}. In LLaVA1.0 \cite{liu2023visual}, a method was proposed to automatically construct a VIF dataset using GPT4, which demonstrated excellent performance in visual question answering (VQA). However, there are two main limitations to the data generated in LLaVA1.0: first, it was constructed using a text-only version of GPT4, which does not accept images as input; and second, it targeted only English.

Subsequently, LLaVA1.5 \cite{liu2023improved} incorporated the multilingual instruction dataset ShareGPT~\cite{sharegpt}, demonstrating its potential in multilingual processing. However, ShareGPT uses an instruction following (IF)~\cite{chen2023visual} dataset for LLMs, still suffers from a lack of vision information. To address this issue, ShareGPT4V~\cite{chen2023sharegpt4v}, a VIF dataset created using GPT4-V, which accepts image information as input, was released. 
ShareGPT4V is also limited because it consists only of English question-answering, posing a constraint in aligning multiple languages to acquire multilingual information.

In this context, we propose constructing a multilingual VIF dataset based on object relational information and a multilingual LMM that efficiently utilizes this dataset. 
The proposed multilingual VIF dataset was composed of 23,496 question-and-answer pairs centered around objects, locations, atmospheres, and conversations to ensure the diversity of expressions. The target languages were selected considering linguistic diversity by choosing English, Chinese, and Korean, which belong to different language families \cite{fitzgerald-etal-2023-massive,park2021klue}.

\newcommand{\cmark}{\ding{51}}%
\newcommand{\xmark}{\ding{55}}%

\begin{table*}[ht!]
    \centering
    \caption{Summary of multi-modal instruction tuning datasets. `Visible’ refers to the including of images in the data generation process. The availability of a `Parallel’ pertains to whether the dataset can be used translation task.}
    \small 
    \begin{adjustbox}{max width=\textwidth}
    \begin{tabular}{@{}l|ccccccccc@{}}
    \toprule
       Dataset & Domain & Data Type & \# of Words  & Visible & Captioned by & \# of Instances & Multilingual& Parallel & Open\\
       \midrule
        MiniGPT4 & Daily life & Description, Discourse & 80 $\sim$ & \xmark & Template-based & 5K & \xmark & \xmark & \cmark\\ 
        MultiInstruct & General & Description, Reasoning & $\sim$ 100 & \xmark & Template-based & $\sim$ 235K  & \xmark & \xmark & \xmark \\  
        InstructBLIP & Daily life & Description, Reasoning, Discourse & $\sim$ 200 & \xmark & Template-based & $\sim$ 1.6M & \xmark & \xmark & \xmark \\ 
        LLaVA & Daily life & Description, Reasoning, Discourse & $\sim$ 200 & \xmark & GPT-based & 1.15M & \xmark & \xmark & \cmark \\ 
        MultiModalGPT & General & Description, Discourse & $\sim$ 200 & \xmark & GPT-based & 6K & \xmark & \xmark & \xmark \\ 
        SharedGPT4V & General & Description, Reasoning, Discourse & $\sim$ 200 & \cmark & GPT-based & 100K & \xmark & \xmark & \cmark\\
        LVIS-INSTRUCT & Daily life & Description & $\sim$ 100 & \cmark & GPT-based & 220K & \xmark & \xmark & \cmark\\
        M$^3$IT & General & Description, Reasoning & $\sim$ 200 & \xmark & GPT-based & 2.4M & \cmark & \xmark & \cmark\\ 
       \midrule
        Ours & Daily life & Description, Discourse & $\sim$ 200 & \cmark & GPT-based & 91K & \cmark & \cmark & \cmark\\
    \bottomrule
    \end{tabular}
    \end{adjustbox}
    \label{tab:dataset_compare}
\end{table*}

We also propose the development of a multilingual LMM, X-LLaVA, utilizing the proposed data. X-LLaVA is a model that enhances LLaVA1.5, by applying the following three enhancement methods: (1) \textbf{vocabulary expansion} for target language, (2) pretraining for \textbf{connecting knowledge} across multiple languages, and (3) \textbf{multilingual VIF}. First, bilingual-based vocabulary expansion involves adding words to a pretrained language model to strengthen the relatively limited vocabulary of Korean compared to English \cite{lu2023ziyavisual, cui2023efficient}. Second, additional pretraining was conducted to link the English and Korean knowledge. Third, we conducted multilingual training using the proposed VIF dataset.

Experimental results showed that the X-LLaVA model demonstrated an average improvement of approximately 5.2\% in three Korean quantitative evaluations compared to the previously proposed KoLLaVA model. In addition, it achieved the highest performance in two out of five English quantitative evaluations. In qualitative evaluations, preference assessments using GPT4-V demonstrated that our model generated responses in both English and Korean that were 19-93\% superior to existing models. Through qualitative analysis, we highlighted that the proposed bilingual training enhanced specific language vocabulary, leading to better performance in writing evaluations. The contributions of this study can be summarized as follows:
\begin{itemize}
\item We propose a training framework of multilingual LMM for enriching a specific language availability
\item We have constructed multilingual VIF dataset based on different task-oriented types 
\item Through an in-depth analysis, we demonstrate the real-world effectiveness of the multilingual approach employed in our dataset.
\end{itemize}

Finally, we emphasize that the 91K datasets and models constructed in this study can be implemented with relatively small resources, costing approximately \$3,200 and utilizing an A6000 GPU.

\section{Related Work}

\subsection{Vision-Language Models}

With the advancement of LLMs, proposals have been made to extend LLMs to include additional modalities~\cite{zhang2023vision}. The primary idea was to focus on aligning information between vision and language \cite{NEURIPS2022_960a172b}. A prime example of this is CLIP~\cite{radford2021learning} and ALBEF~\cite{ALBEF}, which integrated representations of images and text using contrastive learning \cite{pmlr-v119-chen20j, lee-etal-2022-efficient} to unify distinct types of information. Subsequent enhancements, as observed in BLIP~\cite{li2022blip} and BLIP-2~\cite{Li2023BLIP2BL}, utilized assorted data and Q-Former’s trainable query vectors to strengthen this alignment. Most recently, MiniGPT4~\cite{zhu2023minigpt} proposed a fine-tuning method to generate responses that are more aligned with the user intent, demonstrating the potential for conversational image-text models. Concurrently,  InstructionBLIP~\cite{dai2023instructblip}, LLaVA1.0~\cite{liu2023visual}, and LLaVA1.5~\cite{liu2023improved} have advanced our understanding of complex prompts through more sophisticated visual instruction finetuning (VIT) \cite{liu2023visual}.

\subsection{Visual Instruction Following Datasets} 

In LLMs, IF is used to ensure that the language model generates responses that align with user objectives. Recently, there has been a proposal for research to create a VIF dataset that includes image data in the IF. The construction of a VIF dataset is costly and time-consuming because it requires the simultaneous consideration of images, queries, and answers. Therefore, automatic generation methods are commonly used, with two primary approaches: one using GPT for data generation and the other using a template-based method that transforms existing data using predefined templates.

Table~\ref{tab:dataset_compare} presents a comparison of the representative VIF datasets. The initial versions of the VIF dataset were constructed using template-based models. Multi-Instruct \cite{li2023otter} and InstructBLIP, which fall under this category, are fast and cost-effective as they involve rule-based transformation of existing data. However, they have the limitation of being oriented towards specific tasks such as image captioning or classification.

In contrast to template-based construction, LLaVA introduces a more flexible generative data construction method that utilizes the GPT. Using object location and caption information from COCO~\cite{10.1007/978-3-319-10602-1_48}, LLaVA constructed 158K diverse VIF datasets with three different styles: detailed description, complex reasoning, and conversational. However, because these datasets do not use images in their generation, SharedGPT4V~\cite{chen2023sharegpt4v}, and LVIS-INSTRUCT4V~\cite{wang2023instruct4v}, which include images in their construction, were proposed. However, these datasets are predominantly written in a single language. To address the need for multilingual capabilities, the M\textsuperscript{3}IT dataset was released ~\cite{li2023m3it}. M\textsuperscript{3}IT is an instruction-tuning dataset comprising 40 tasks translated into 80 languages that offers broad accessibility.

\section{Data Generation} \label{sec:data_generation}

\begin{figure}
  \includegraphics[width=\linewidth]{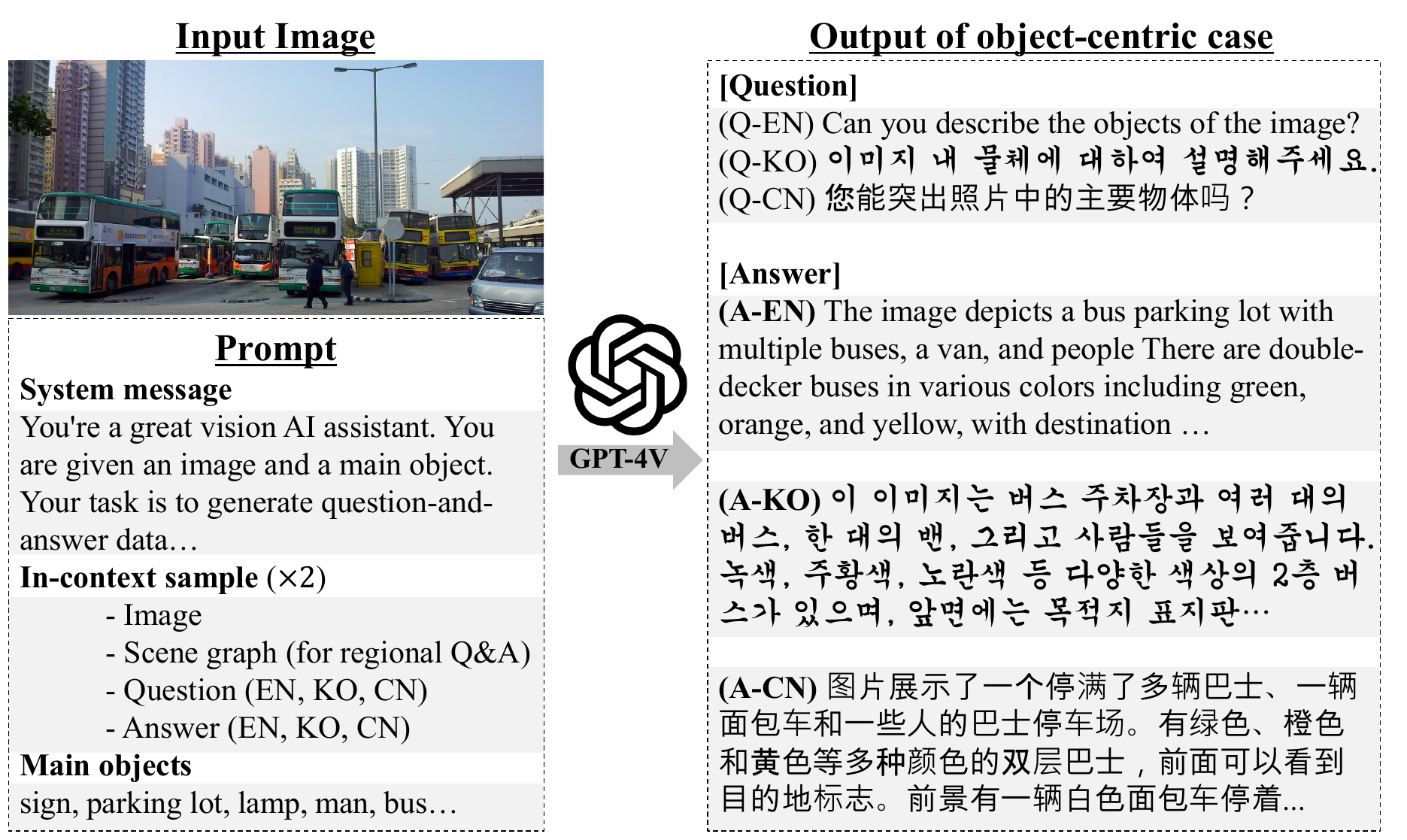}
  \captionsetup{font=small, skip=2pt}
  \caption{An example of prompt and result using data construction.}
  \label{fig:prompt_example}
\end{figure}
\vspace{-0.2cm}

In this study, we were inspired by the VIF data generation method using the GPT of LLaVA and have built upon it. However, to minimize the loss of information from the images and include more detailed information, we directly input the image and object information into the GPT4-V model to construct our data. 
We constructed four types of multilingual VIF datasets (\texttt{mvif}) for three languages (English, Korean, and Chinese): (1) Object-centric, (2) Location-centric, (3) Atmosphere-centric, and (4) Conversation.

\subsection{The Focus of Data Building}

The \texttt{mvif} data proposed in this research concentrate on the relational factual information between objects. This focus diverges from the description and reasoning-centered question-answering proposed by LLaVA, leading to minimal information redundancy between the two datasets. Although LLaVA’s data are commendable, we assessed whether data designed for reasoning purposes might incorporate subjective viewpoints, thereby potentially introducing bias toward certain objects. Therefore, our study aims to develop a functional-relationship-based multilingual VIF dataset that, deliberately avoids overlap with LLaVA.

The target languages selected were English, Chinese, and Korean, each belonging to a distinct language family. This choice was intended to evaluate how multilingual training affects the languages of different cultures and character systems.

\subsection{Image Selection Criteria}
To construct the \texttt{mvif} dataset, 23,496 images from the visual Genome \cite{krishna2017visual} were used. A challenge was encountered when generating data using GPT4: if an image contained fewer than three major objects, the constrained context could limit the diversity of question answers. However, answering questions generated using images with over ten objects often results in a focus on objects that are either exceedingly small or insignificant. Consequently, we speculate that images selected from the visual Genome, where the number of main objects corresponds to  $3\le m\le 10$.

\subsection{Proposed VIF Dataset} \label{sec:VIF_dataset}
Figure~\ref{fig:prompt_example} shows an example of the method used to construct the proposed \texttt{mvif} dataset. As illustrated, an image and a prompt, which are metadata for question generation, were fed into GPT4-V. Subsequently, GPT4-V was designed to generate questions and answers in three languages. For conversation data, we designed a prompt to produce eight pairs of dialogues for each image in a multi-turn format. For the dataset construction, we provided two seed examples to GPT4-V to guide the construction of data suitable for the purpose through in-context learning.
A total of \$3,200 was used to generate 91K data points. Detailed prompts used in data construction; the four types of generated data samples and inspection procedure can be found in the Appendix G.\\

\noindent\textbf{(1) Object-centric image description.}\ 

Object-centric data focuses on providing detailed description of objects in an image, comprising questions and answers that include the shape, condition, and characteristics of the objects. The aim of constructing these data was to facilitate the learning of the intimate details of images by focusing on the specific attributes of the objects as they appear. Additionally, as shown in the ``Main objects'' section of Figure~\ref{fig:prompt_example}, a list of main objects was inputted into the GPT4-V prompt to prevent errors in object specification that might occur during question generation.\\

\noindent\textbf{(2) Location-centric image description.}\ 
Location-centric data is a type of question-answering data that focuses on describing the relative positions of objects within an image. However, when the same object appears multiple times in an image, this perspective can alter the location information. To address this effectively, we enabled GPT4-V to autonomously generate a relationship graph that served as the basis for answering the question. Consequently, when GPT4-V receives an image and a list of objects, it first generates a scene graph and then produces locational questions and answers regarding the image. \\

\noindent\textbf{(3) Atmosphere-centric image description.}\ 

Atmosphere-centric data include descriptions that focus more on the overall ambiance of an image than on individual objects. It encompasses a holistic depiction of the complex interplay among multiple objects. \\

\noindent\textbf{(4) Conversational question and answering} \
Conversational data is structured as an 8-turn Q\&A dataset to incorporate more in-depth and extensive information regarding the images. Unlike other datasets, this dataset is designed to infer human emotions or include subjective information about the mood of the image.

\section{Proposed Multilingual Model} \label{sec:Proposed_model}
\begin{figure}
  \includegraphics[width=\linewidth]{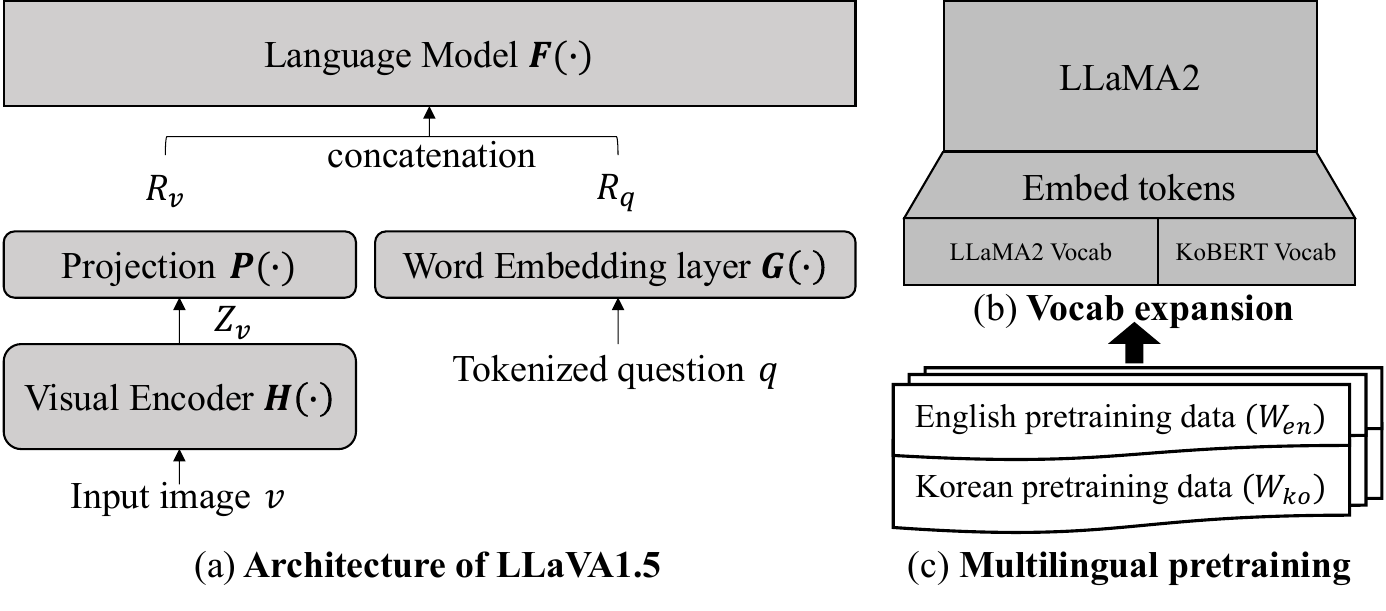}
  \captionsetup{font=small, skip=2pt}
  \caption{(a) Architecture of LLaVA1.5  \& (b,c) The proposed language model pretraining}
  \label{fig:LLaVA_arch}
\end{figure}

In this section, we introduce the proposed X-LLaVA model, an effective approach for multilingual processing through multilingual VIT~\cite{liu2023visual}. X-LLaVA applies the following three enhancement methods to the same model structure as LLaVA1.5: (1) vocabulary expansion for the target language, (2) pretraining for multilingual knowledge association, and (3) multilingual VIT. Figure~\ref{fig:LLaVA_arch} demonstrates the three proposed methods and the structure of LLaVA1.5. 
\vspace{-0.2cm}

\subsection{Recap of LLaVA1.5}
Figure~\ref{fig:LLaVA_arch} (a) shows the basic structure of the LLaVA1.5 model. LLaVA1.5 basically consists of a visual encoder and an LLM for natural language generation. The visual encoder utilizes a pretrained CLIP's Vision Transformer~\cite{yuan2021tokens} $H(\cdot)$, and the LLM $F(\cdot)$ utilized the pretrained LLaMA2-based models~~\cite{touvron2023llama,peng2023instruction}. LLaVA uses image $v$ and query $q$ as inputs. In the case of image $v$, the output representation from the visual encoder, $H(v) = Z_v \in \mathbb{R}^{576\times 1024}$, is converted into a vision-language representation $R_v \in \mathbb{R}^{576\times5120}$ through a projection layer $P(\cdot) : \mathbb{R}^{1024} \to \mathbb{R}^{5120}$. For text $q$, it passes through the embedding layer $G(\cdot)$ of LLaMA to generate the text representation $G(q) = R_q \in \mathbb{R}^{(|q|, 5120)}$. $R_q$ and $R_v$, generate through these two processes are concatenated and then passed through the entire layer of the LLaMA2 to produce a response. In this context, the projection layer serves the function of transforms image representation \( Z_v \) into a word embedding format that can be understood using the LLaMA2.

To achieve image-language alignment, we train the process to connect the two representations, which LLaVA does in two steps. The first is image-text alignment through image captioning, and the second is VIT. X-LLaVA is trained in the same manner, and the details of the two phases are described in Section~\ref{sec:x_llava}.

\subsection{Enriching the LLM Vocabulary} \label{sec:enriching_voca}
In the LLaVA model, when querying in Korean for the LLaMA2-13B language model, issues arise, such as responses in English or English-Korean code-switching. This stems from a problem with the tokenizer, where 89.7\% is in Latin script, while Korean only constitutes 0.37\%, leading to insufficient Korean expressiveness and biases in the pretraining data owing to lexical bias. To address these issues, we expanded the Korean vocabulary in the LLaMA2 and conducted additional pretraining for knowledge infusion. (Figure~\ref{fig:LLaVA_arch} (b), (c))

Vocabulary expansion involves adding 7,478 words from the KoBERT\footnote{https://github.com/SKTBrain/KoBERT} vocabulary to the LLaMA2 tokenizer. And we randomly initialize embeddings for these newly added words. Ultimately, the proposed tokenizer possessed a dictionary of 39,478 entries. As a subsequent step, the model was further enhanced with knowledge information using English Wikipedia data $\text{W}_{en}$ and Korean Wikipedia data $\text{W}_{ko}$. Through this process, our model learns representations for the newly added vocabulary. If the pretraining dataset (7.8GB) is defined as $D_{pt} = \{\text{W}_{en}, \text{W}_{ko}\}$, then the loss function $\mathcal{L}_{PT}(\cdot)$ is expressed as follows.

\begin{equation}
\mathcal{L}_{PT}(\theta) = -\sum_{i}^{|D_{pt}|}\sum_{j}^{|x_i|} \log P(x_{i, j} | x_{i, <j};\theta)
\end{equation}

Here, $|D_{pt}|$ is the size of $D_{pt}$, $|x_i|$ denotes the number of tokens in $i$-th data sample $x_i$. $x_{i, j}$ represents $j$-th token of sequence $x_i$, and $x_{i, <j}$ represents the sequence of tokens before the $j$-th token. In this context, $\mathcal{L}_{PT}(\theta)$ is the causal language modeling loss function, where $\theta$ denotes the model parameters.

\subsection{X-LLaVA} \label{sec:x_llava}

In this section, we describe the method for training X-LLaVA using the LLaMA2 model, which has proceeded word expansion and bilingual dictionary pretraining, as previously introduced X-LLaVA, like LLaVA, is trained in two stages: image-language connection via captioning and multilingual VIT. However, unlike LLaVA1.5, to efficiently conduct multilingual training, we follow the cross-lingual language model pretraining method ~\cite{conneau2019cross}, simultaneously utilizing a mix of English and Korean for training.

In the first stage, we train only the projection layer $P(\cdot)$ using the image-caption datasets LLaVA-CC3M~\cite{liu2023visual} $(C_{en})$ and its machine-translated Korean counterpart, LLaVA-KoCC3M$(C_{ko})$. This stage involves representation learning in which image representations are converted into word embeddings that are comprehensible to the LLaMA2. During this process, both Korean and English are learned concurrently while simultaneously aligning [image-English-Korean]. We define the dataset for Stage-1 as $D_{s1} = \{C_{en}, C_{ko}\}$.

In the second stage, we conducted VIT on X-LLaVA to enhance its capabilities as a multilingual visual assistant. For VIT as described in ~\cite{liu2023visual}, we use the LLaVA instruct dataset (158K, $ L_{en}$), its machine-translated counterpart (158K, $L_{ko}$), and the \texttt{mvif} dataset (91K, $L_{our}$) generated in Section~\ref{sec:data_generation}. In this stage, unlike the first stage, we train the projection layer and language model simultaneously. Define the dataset for Stage-2 training as $D_{s2} = \{L_{en}, L_{ko}, L_{our}\}$. The formula for training the Stage-2 can be expressed as follows:
\begin{equation} 
\mathcal{L}_{s}(\theta)\!= -\sum_{i}^{|D_{s}|}\sum_t^T\sum_j^{|a_i^{(t)}|}\log P(a_{i, j}^{(t)}|X_{i, <j}^{(t)} ; \theta)
\end{equation}
Where $X_{i, <j}^{(t)} = \{v_i, q_i^{(1)}, a_{i}^{(1)}, \cdots, q_{i}^{(t)}, a_{i, <j}^{(t)}\}$, $T$ represents the total number of conversation turns. In Stage 1, $T=1$ because the dataset $D_{s1}$ is composed of a single turn. In Stage 2, $T=1$ is also true in all case, except for multi-turn conversations.

In the dataset $D_s$, which can be either $D_{s1}$ or $D_{s2}$ depending on the stage, $v_i$, $q_i^{(t)}$, and $a_i^{(t)}$ denote the $i$-th component of the image, the question (instruction) in turn $t$, and the answer in turn $t$, respectively.

\section{Quantitative Evaluation} \label{sec:experiment}

In this section, we describe the quantitative evaluation methods and criteria for the proposed X-LLaVA. Through these comparisons, we aim to address the three research questions proposed in Section~\ref{sec:introduction}: (1) What impact does vocabulary expansion, intended to enhance multilinguality, have on vision-language models? and (2) How does bilingual training affect the relationship between these two languages? and (3) Which aspects of the model were strengthened by utilizing our proposed \texttt{mvif} data?

\subsection{Experiment Environments}

To ensure a fair comparison of LMMs, we must define task selection for evaluation and specify the LMM model used for evaluation. Below are the benchmark datasets used for evaluation, with the following characteristics for each benchmark:

\begin{itemize}
    \item{\textbf{(English)}} \textbf{VQA2.0:} A dataset containing open-ended questions about images~\cite{goyal2017making}, \textbf{GQA:} A VQA-format dataset considered Scene Graph~\cite{hudson2019gqa}, \textbf{LV} (LLaVA\textsuperscript{w} from \cite{liu2023visual}) and \textbf{POPE} \cite{Li-hallucination-2023}
    \item{\textbf{(Korean)}} \textbf{KoViz:} A VQA-format dataset and \textbf{KoLiv:} A VQA-format dataset considered Korean culture and daily life \cite{kim2019korean}
    \item{\textbf{(English-Korean)}} \textbf{BVQA}~\cite{kim2024bok}: A VQA dataset considering \textbf{B}ilingual Out-side Knowledge
\end{itemize}

For our experiments, we converted the VQA2.0 and BVQA~\cite{kim2024bok} datasets into the VIF format using the VQA-to-VIF data transformation method proposed in LLaVA1.5. Following this conversion, we proceeded with VIT over all the training sets from the proposed benchmark in only one epoch. The evaluation methodology and prompts were adopted directly as proposed in LLaVA1.5 (See Appendix C). Experimental environments and answers generated for each model were made publicly accessible\footnote{github.com\//AnonymousMercy\//NACCL\_submit} to ensure reproducibility and facilitate comparison of the models.

\subsection{Intrinsic Evaluation of X-LLaVA}

\begin{table}[]
    \small
    \begin{tabular}{ll|lcccc}
    \toprule 
    Model           & VIF                 & BVQA$^{k}$    & BVQA$^{e}$   & GQA    \\
    \midrule
    XLLaVA(-V,-P)    &                    & 51.5              & 33.0   & 62.3      \\
                    & + O                 & 51.9              & \textbf{36.0}   & 61.9      \\                                        
    \midrule
    XLLaVA(-P)      &                     & 56.4           & 32.0   & 62.1   \\
                    & + O                 & 56.6           & 32.3   & 62.5  \\
    \midrule
    XLLaVA          &                     & 57.6           & 33.5          & 63.3  \\
                    & + O                 & \textbf{57.9}  & 34.3 & \textbf{64.0}\\

    \bottomrule
    \end{tabular}
    \caption{Intrinsic evaluation. Where (-V) represents without vocabulary expansion, and (-P) denotes without multilingual pretraining step. Metric is Accuracy(\%).}
    \label{tab:intrinsic_eval}
\end{table}

An intrinsic evaluation was conducted to explore the three research questions we proposed. To achieve this, we train the three models under different conditions. Table~\ref{tab:intrinsic_eval} lists the training environments and performances of the three models. X-LLaVA refers to the model that underwent both vocabulary expansion and knowledge enhancement (\ref{sec:enriching_voca}) as well as the VIT (\ref{sec:x_llava}) proposed in Section~\ref{sec:Proposed_model}. X-LLaVA(-P) is a model created to compare the effects of pretraining methods on Koreans and English data proposed in Section~\ref{sec:enriching_voca}. This model is a version of X-LLaVA that does not utilize Wiki for \textbf{p}retraining during its training phase. X-LLaVA(-V,-P) represents a model that neither underwent \textbf{v}ocabulary expansion nor used Wiki for \textbf{p}retraining, essentially using pure LLaMA2. Finally, to assess the impact of the \texttt{mvif} data proposed in Section~\ref{sec:data_generation}, we compared the results of each model with and without the addition of \texttt{mvif}.\\

\noindent \textbf{The influence of Enriching Vocabulary.} 

Comparing the X-LLaVA and X-LLaVA(-V,-P) models in Table~\ref{tab:intrinsic_eval}, we observe an average of 6.1 points for Korean and 0.8 points for English. Therefore, the vocabulary expansion and pretraining proposed in Section~\ref{sec:enriching_voca} not only significantly improves the Korean performance of the model with expanded vocabulary but also enhances the performance of the existing English model. \\

\noindent \textbf{The influence of Pretraining.} A comparison between the X-LLaVA and X-LLaVA(-P) models showed that additional pretraining using Wikipedia uniformly enhanced the performance in both Korean and English, with a particularly notable improvement in Korean. Therefore, the effectiveness of pretraining in Korean and English using Wikipedia was evident. \\

\begin{table*}
\small

\begin{tabular}{llll|ccc|ccccc}
\toprule

LMM                 & LLM         & \#PT   & \#VIT       & BVQA\textsuperscript{k}    & KoViz   & KoLiv & BVQA\textsuperscript{e}   & VQA & GQA  & LV & POPE\\
\midrule
BLIP-2         & Vicuna13B  & 129M  & -        & -         & -            & -     & -    & 41            & 41   & -   & 85.3   \\
InstructBLIP   & Vicuna7B   & 129M  & 1.2M     & -         & -            & -     & -    & -             & 49.2 & -     & -    \\
InstructBLIP   & Vicuna13B  & 129M  & 1.2M     & -         & -            & -     & -    & -             & 49.5 & -   & 78.9   \\
LLaVA1.5       & Vicuna7B   & 558K  & 665K     & 16.2      & 33.9         & 44.9  & 25.1 & 78.5          & 62.0 & 64.7     & \textbf{85.9} \\
LLaVA1.5       & Vicuna13B  & 558K  & 665K     & 27.9      & 24.4         & 33.4  & 26.1 & \textbf{80.0} & 63.3 & 65.7  & \textbf{85.9} \\
LLaVA1.5(O)    & Vicuna13B  & 558K  & 756K     & 32.6      & 24.6         & 23.2  & 29.1 & 78.1          & 45.3 & \textbf{70.4}  & 85.8  \\
LLaVA1.5(B)    & Vicuna13B  & 558K  & 857K     & 54.5      & 50.3         & 52.1     & 33.5 & 76.4           & 63.0    & 22.8  & 85.8    \\
KoLLaVA        & Synatra7B  & 595K  & 612k     & 45.3      & \textbf{55.9}& 54.2  & 5.5  & -             & -    & -  &  -  \\
\midrule
X-LLaVA         & Ours       & 1.2M  & 407K  & \textbf{57.9}& 51.3 & \textbf{61.7} & \textbf{34.3} & 75.5 & \textbf{64.0} & 57.5 & 85.5 \\

\bottomrule
\end{tabular}
\caption{Extrinsic evaluation results. Where (O), (B) represents training with \texttt{mvif} and BVQA dataset,\texttt{\#PT} is the number of pretraining data, \texttt{\#VIT} is the number of VIT data. POPE is a benchmark for evaluation of hallucination.}
\label{tab:extrinsic_eval}
\end{table*}

\noindent \textbf{The influence of VIT using \texttt{mvif}.} 
When models were tuned with the proposed dataset (+O), a performance improvement ranging from 0.2 to 3 was observed across almost models for the target language. Although the extent of improvement is modest, it is noteworthy that despite the grammatical differences between Korean and English, where knowledge loss might be anticipated, there was an observable enhancement in the English performance. This indicates that multilingual VIF can be expected to improve performance in both less- and high-resource languages.

\subsection{Extrinsic Evaluation of X-LLaVA}

We conducted a comparative evaluation of the performance of our X-LLaVA model in Korean and English against other LMMs. The models compared were BLIP-2, InstructBLIP, LLaVA1.5, and KoLLaVA, and the distinctive features of each model are presented in Table~\ref{tab:extrinsic_eval}. \\

\noindent \textbf{Overall.} In the Korean evaluation (BVQA\textsuperscript{k},Koviz, and KoLiv) presented in Table~\ref{tab:extrinsic_eval}, X-LLaVA demonstrated significantly higher performance, scoring on average 57.0 points. Interestingly, in the case of English (VQA, GQA, BVQA\textsuperscript{e}, LV, POPE), X-LLaVA also showed the highest performance in BVQA\textsuperscript{e} and GQA. \\

\noindent \textbf{The effect of multilingual training.} 

Typically, when training languages with different character systems, the performance of a relatively highly resourced language may deteriorate~\cite{pires-etal-2019-multilingual}. However, when the multilingual training methods and data (\texttt{mvif}) we proposed, no decrease in performance was observed. When comparing the English BVQA\textsuperscript{e} and GQA scores of LLaVA1.5 and X-LLaVA, they showed 8.2 and 0.7 points higher performance, respectively. However, for VQA2.0, LLaVA1.5’s performance was 4.5 points higher. During analysis, we observed that X-LLaVA generally performed better on GQA and BVQA, which asked about relationships and knowledge. \\

\noindent \textbf{Comparison of X-LLaVA with KoLLaVA.} KoLLaVA\footnote{github.com\//tabtoyou\//KoLLaVA} is the Korean version of LLaVA1.5, a model trained after automatically translating CC3M, VQA2.0, GQA, and Visual Genome data used in LLaVA1.5. Additionally, it was trained using the Korean version of the BVQA. However, as only the 7B model is currently publicly available, it may be challenging were used to evaluate the same levels. However, the published LLaVA1.5 13B model shows an average of 0.96 points higher in english than that of the 7B model, X-LLaVA demonstrates a 5.2 point higher result in korean than KoLLaVA. \\

\noindent \textbf{Comparison X-LLaVA with LLaVA1.5(O or B).} 
LLaVA1.5 was trained on about 1.5 times more data (665K VIFs) then X-LLaVA. Nevertheless, BVQA data has never been utilized for training, which may be disadvantageous for the BVQA evaluation. We trained LLaVA1.5 on Korean and English data for three 3 epochs to tune the BVQA for a fair evaluation. LLaVA1.5(B) in Table~\ref{tab:extrinsic_eval} shows the results of the model tuned using the BVQA data. The results show a significant improvement in Korean performance on the BVQA. On the other hand, this model, being biased towards VQA data, showed lower performance in the writing evaluation (LV). Conversely, LLaVA1.5(O) in Table~\ref{tab:extrinsic_eval}, a model trained on the LLaVA1.5 with \texttt{mvif} data, exhibited the highest performance on LV.

\section{Qualitative Evaluation}
In this section, we describe the qualitative evaluation methods and the results for X-LLaVA. In contrast to quantitative evaluations, which are similar to classification assessments, qualitative evaluations, such as writing evaluations, differ significantly. Although human evaluation may be the fairest approach to qualitative assessments, it is practically challenging. Therefore, in LIMA~\cite{zhou2023lima}, a GPT preference evaluation method that closely resembles human evaluation results was proposed.

In our study, we directly employed the GPT preference evaluation method. The process is as follows: First, we input an image and a question into two models being compared to obtain answers A and B. Then, we provided GPT4 with the image, question, and both answers to receive feedback such as `Answer A is better’, `Answer B is better’, or `Both answers are similar’, and measured the proportions. To compare the standing and generation abilities of recent LMMs in vision language, we used the GPT evaluation dataset proposed by LLaVA\footnote{`qa90\_gpt4\_answer' at github.com\//haotian-liu\//LLaVA}. However, because this dataset is in English, we translated it into Korean, followed by a review from five annotators to ensure data quality. Afterward, we proceeded with the evaluations.

\subsection{Preference Evaluation using GPT4-V}

\begin{figure}[h]
  \includegraphics[width=\linewidth]{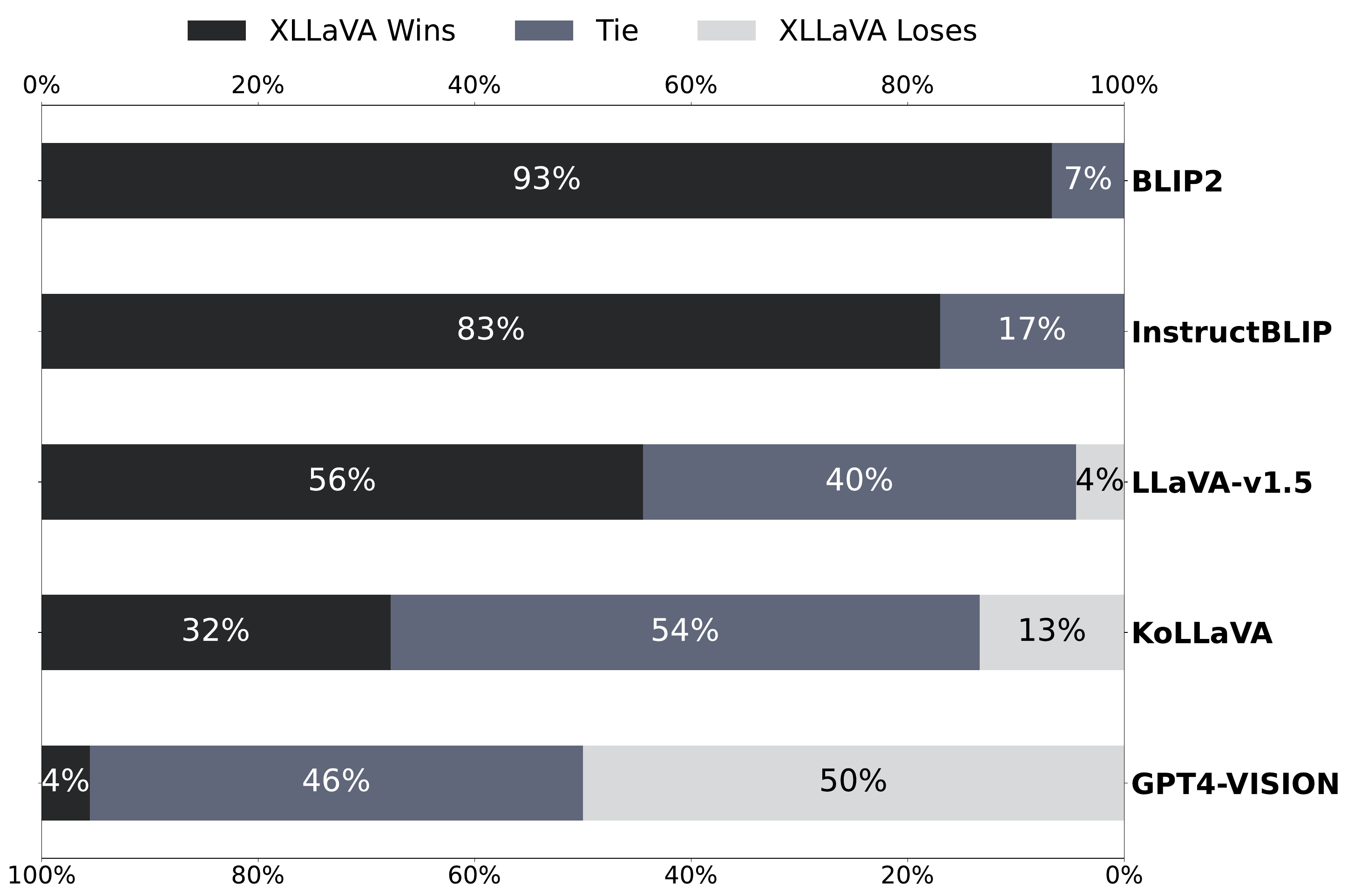}
  \captionsetup{font=small, skip=2pt}
  \caption{Korean Preference evaluation results by GPT4-V}
  \label{fig:qualitative Evaluation (KO)}
\end{figure}

\begin{figure}[h]
  \includegraphics[width=\linewidth]{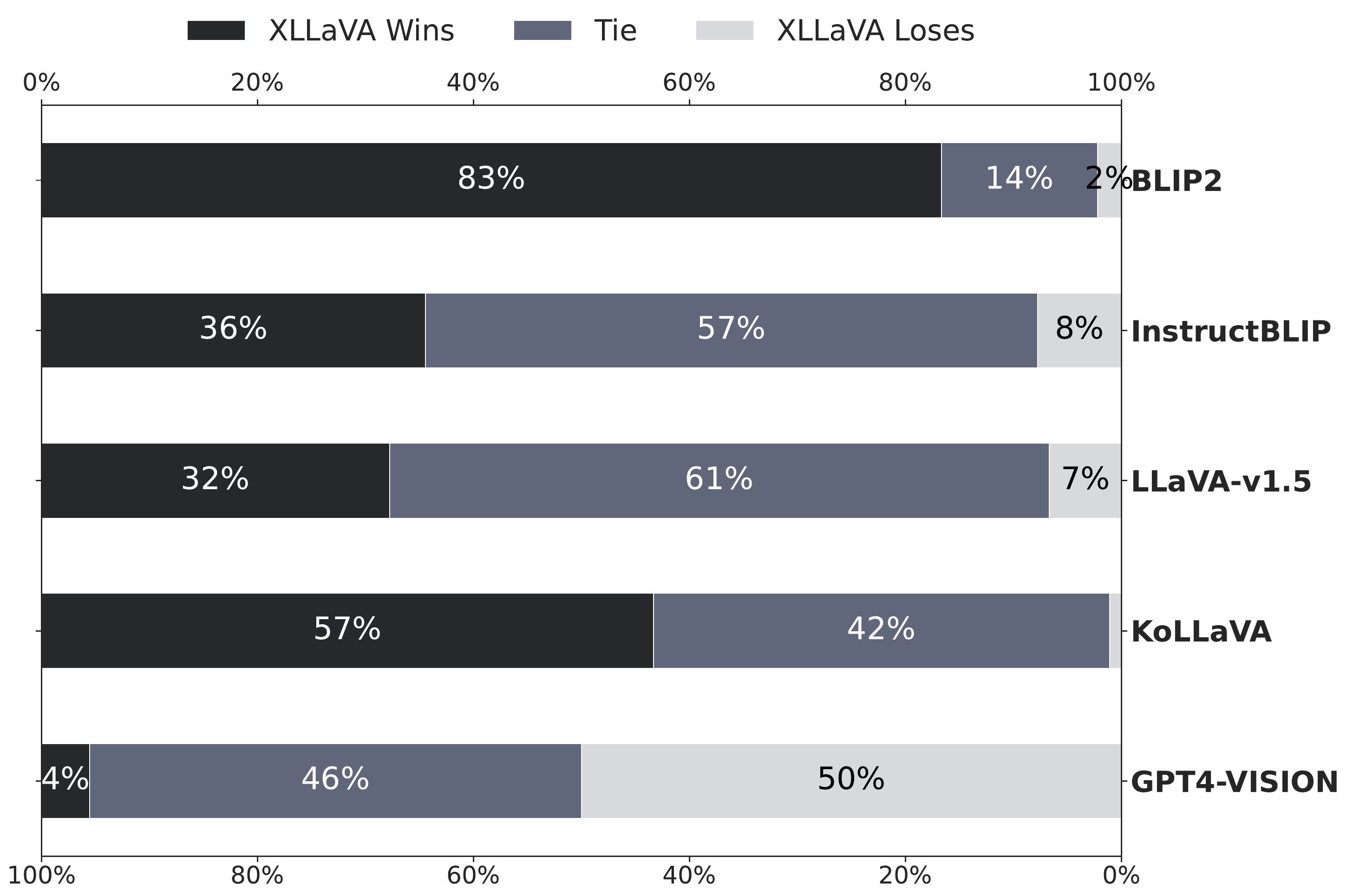}
  \captionsetup{font=small, skip=2pt}
  \caption{English Preference evaluation results by GPT4-V}
  \label{fig:qualitative Evaluation (EN)}
\end{figure}
\vspace{-0.2cm}

\noindent \textbf{Comparing X-LLaVA with others in Korean.} 
Figure~\ref{fig:qualitative Evaluation (KO)} presents the results of the GPT preference evaluation for each model. The X-LLaVA model outperformed all other models, except for the GPT4-V model. Notably, it obtained a 19\% higher preference rate than the KoLLaVA, indicating the exceptional effectiveness of the proposed methods and datasets in enhancing Korean writing skills. \\ 

\noindent \textbf{Comparing X-LLaVA with Others in English.} Figure~\ref{fig:qualitative Evaluation (EN)} shows the results of English GPT preference evaluations. Interestingly, similar to Korean, the X-LLaVA received approximately 25\% higher preference scores for English than LLaVA1.5. This indicates that pretraining of our proposed LLM and \texttt{mvif} datasets can also enhance English writing abilities. \\

\begin{figure}[t]
  \includegraphics[width=\linewidth]{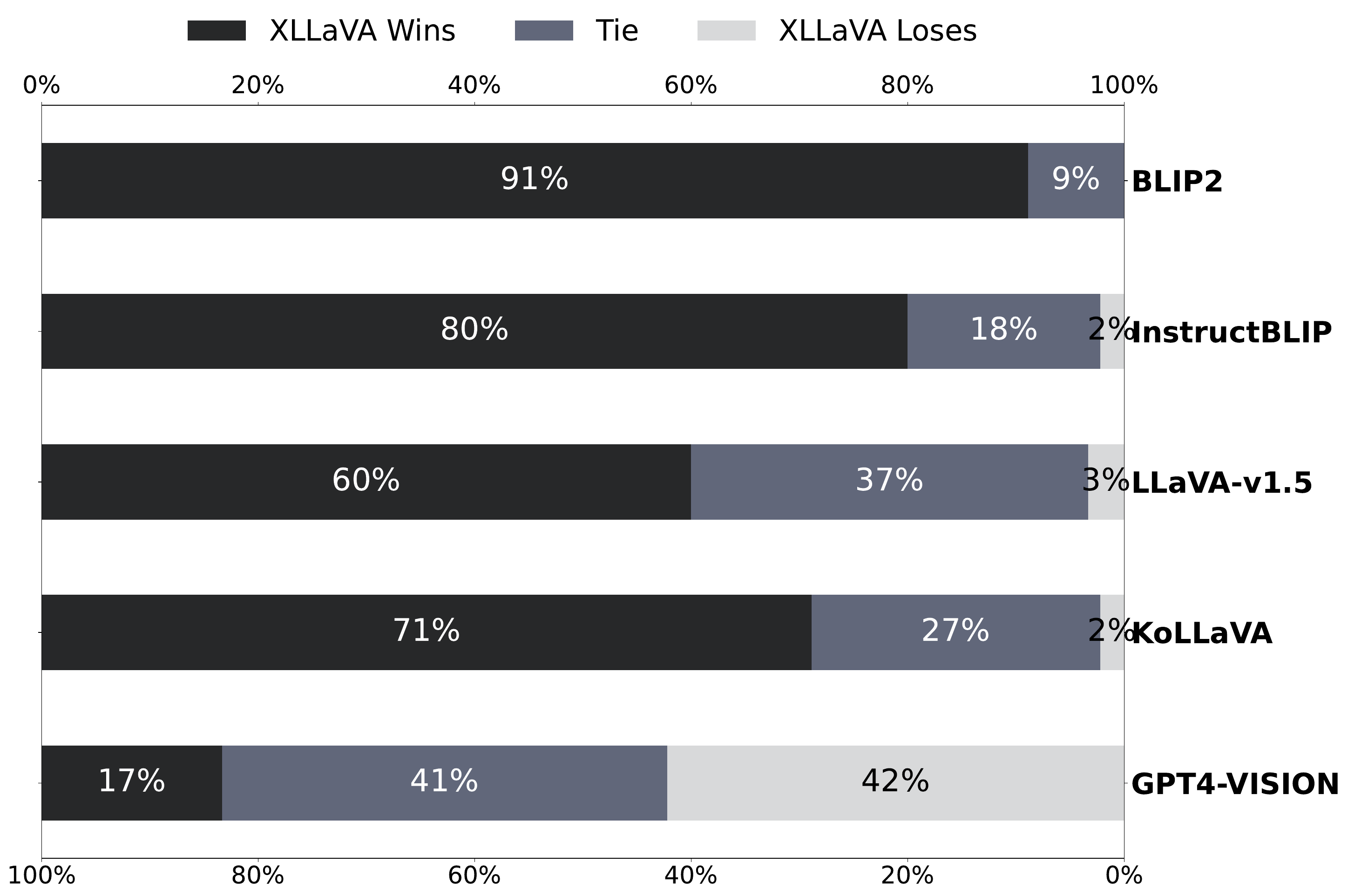}
  \captionsetup{font=small, skip=2pt}
  \caption{Korean Preference evaluation results by GPT4-V when limited to 30 Words.}
  \label{fig:qualitative Evaluation (KO, short)}
\end{figure}

\begin{figure}[t]
\centering
\includegraphics[width=\linewidth]{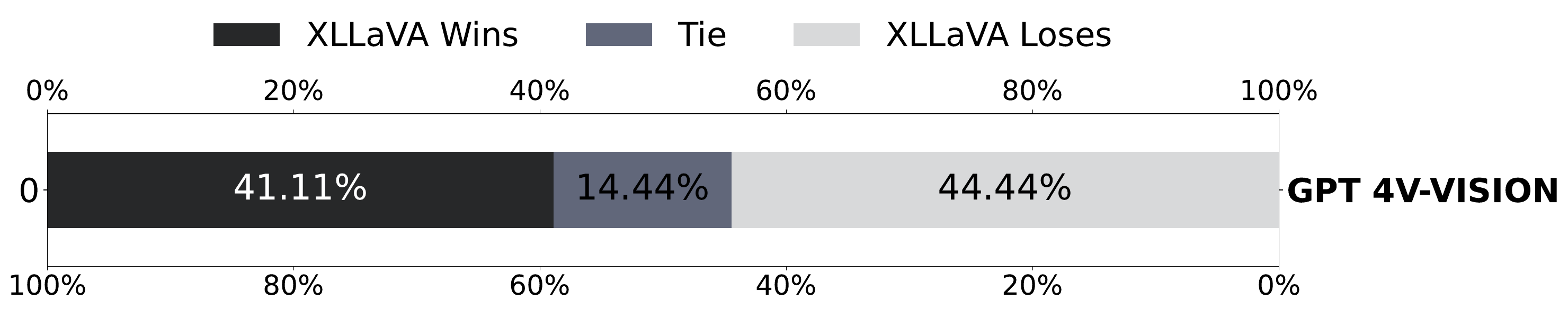}
\caption{Preference evaluation results by human}
\label{fig:kor_llm_eval_human}
\end{figure}

\noindent \textbf{X-LLaVA vs GPT4-V.} Therefore, does evaluator GPT4-V generate better answers than X-LLaVA? We conducted the evaluations by comparing the GPT4-V and X-LLaVA models. Experimental results show that for both languages, GPT4-V’s answers are preferred over those of X-LLaVA, with a significant performance difference. However, these results stem from GPT4-V generating answers that are more than 30\% longer and more verbose compared to LLaVA-based models. This may also be because the GPT rates its own generated content more favorably as it becomes more familiar with it. To mitigate this, in experiments where the answers were limited to 30 words, the results changed significantly, with GPT scoring 42 compared to 17 for X-LLaVA. Detailed statistical analysis related to this can be found in Figure~\ref{fig:qualitative Evaluation (KO, short)} and Appendix E. 

\subsection{Human-assisted Preference Evaluation}

As previously described, the performance of GPT preference evaluation may vary according to the number of words. Consequently, a question arises: Can LIMA's assertion that GPT evaluations are akin to human assessments be extended to the vision-language model proposed in this study? We conducted a human preference evaluation using three human annotators. Figure~\ref{fig:kor_llm_eval_human} presents the results of the human evaluation for GPT4-V and X-LLaVA in the comparative assessment, with the response length restricted to 30 words. Although GPT maintained a slight advantage, the preference scores were nearly identical. However, we observed that GPT evaluations resulted in ties 2.9 times more frequently than human evaluations. This observation can be interpreted to suggest that GPT tends to avoid ambiguous decisions compared to humans, who possess relatively clear criteria. Thus, the vision-language model can be considered as augmenting rather than substituting human evaluations. Details supporting this, along with comprehensive human evaluation results and analyses for the entire model, are available in Appendix F.

\section{Conclusion}

In this study, we propose a framework for constructing data and training models for the efficient multilingual expansion of LMM. For data construction, we suggested a method to easily build multilingual VIF dataset based on the relational metadata between images and objects using GPT4-V. We also demonstrated a framework for efficient multilingual learning, which includes vocabulary enhancement, knowledge reinforcement based on pretraining, and a multilingual VIT framework. The experimental results confirmed that the proposed X-LLaVA model exhibited similar or superior performance compared to existing models that primarily focused on Korean and English as single languages. Finally, our proposed multilingual expansion framework can be trained in 7.5 days with a single A6000 GPU, and the 91K training data can be managed with relatively minimal resources, costing around \$3,200.

\section*{Limitations}
The ultimate goal of this research is to create a multilingual Large Multimodal Model (LMM). However, in this study, we first conducted pretraining in Korean-English and then proceeded with multilingual visual instruction following in Korean-English-Chinese. Consequently, as the Chinese component of the model did not undergo word expansion, it more closely resembles a Korean-English bilingual enhanced model. Therefore, there is a need for further investigation and research into models that have undergone vocabulary enhancement and knowledge connection for more than three languages. An additional factor was the difficulty in finding publicly available Chinese VQA evaluation data, which hindered diverse assessments.
\section*{Acknowledgements}
This research was supported by the National Research Foundation of Korea (2021R1F1A1063474) for KyungTae Lim and Institute of Information \& communications Technology Planning \& Evaluation (IITP) by the Korea government(MSIT) (2022-0-00078, Explainable Logical Reasoning for Medical Knowledge Generation). This research used datasets from The Open AI Dataset Project (AI-Hub) (No. 2022-데이터-위41, 2023-지능데이터-위93).
\bibliography{anthology,custom}
\appendix

\newpage

\onecolumn
\section{Data Generation Example}
\begin{figure}[hbt!]
    \centering
\includegraphics[width=\linewidth]{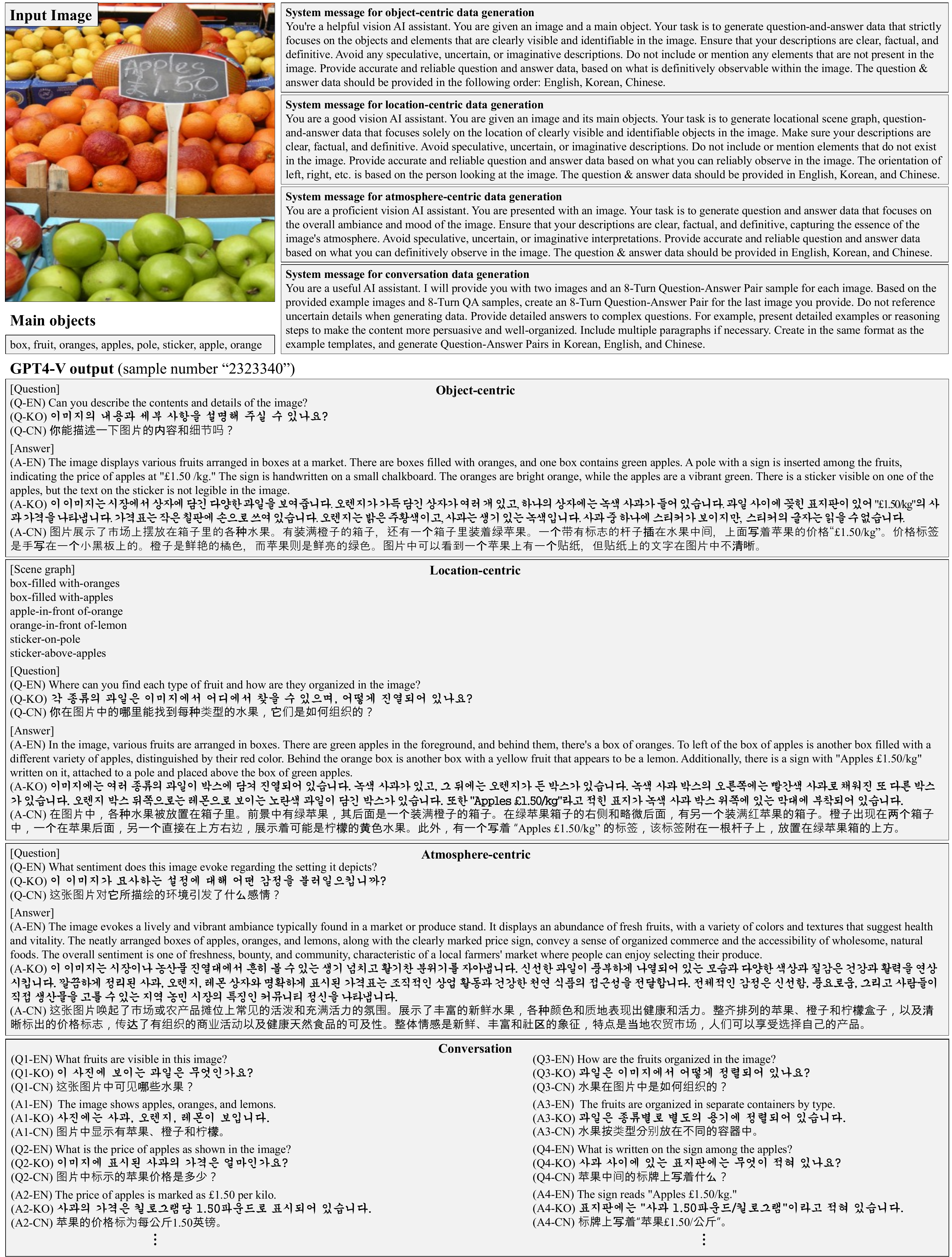}
    \scriptsize\caption{
    Example for Query Generation. An input image, system message, and main objects are given as inputs, and as output, four different query-response samples are generated. EN : English, KO : Korean, CN : Chinese.
  }
    \label{fig:pieplot1}
\end{figure}

\newpage

\twocolumn

\section{Data Statistics}

\begin{figure}[hbt!]
\centering
    \includegraphics[width=1.05\linewidth]{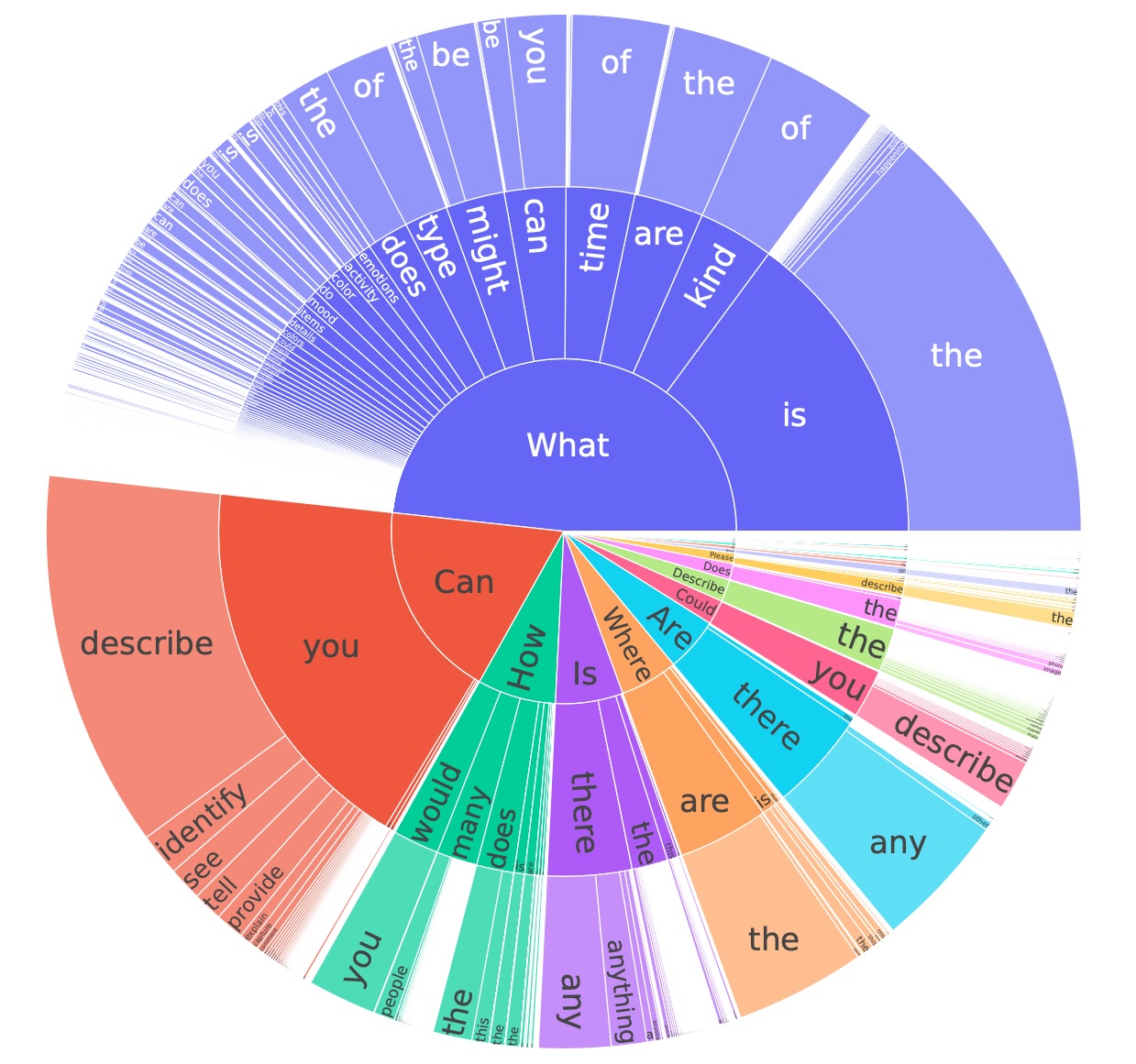}
    \captionsetup{format=plain}
    \caption{
This chart displays the frequency of words found in the \texttt{mvif} questions, organized according to their syntactic order.
  }
    \label{fig:pieplot1}
\end{figure}

\begin{figure}[hbt!]
\centering
  
    \includegraphics[width=1.05\linewidth]{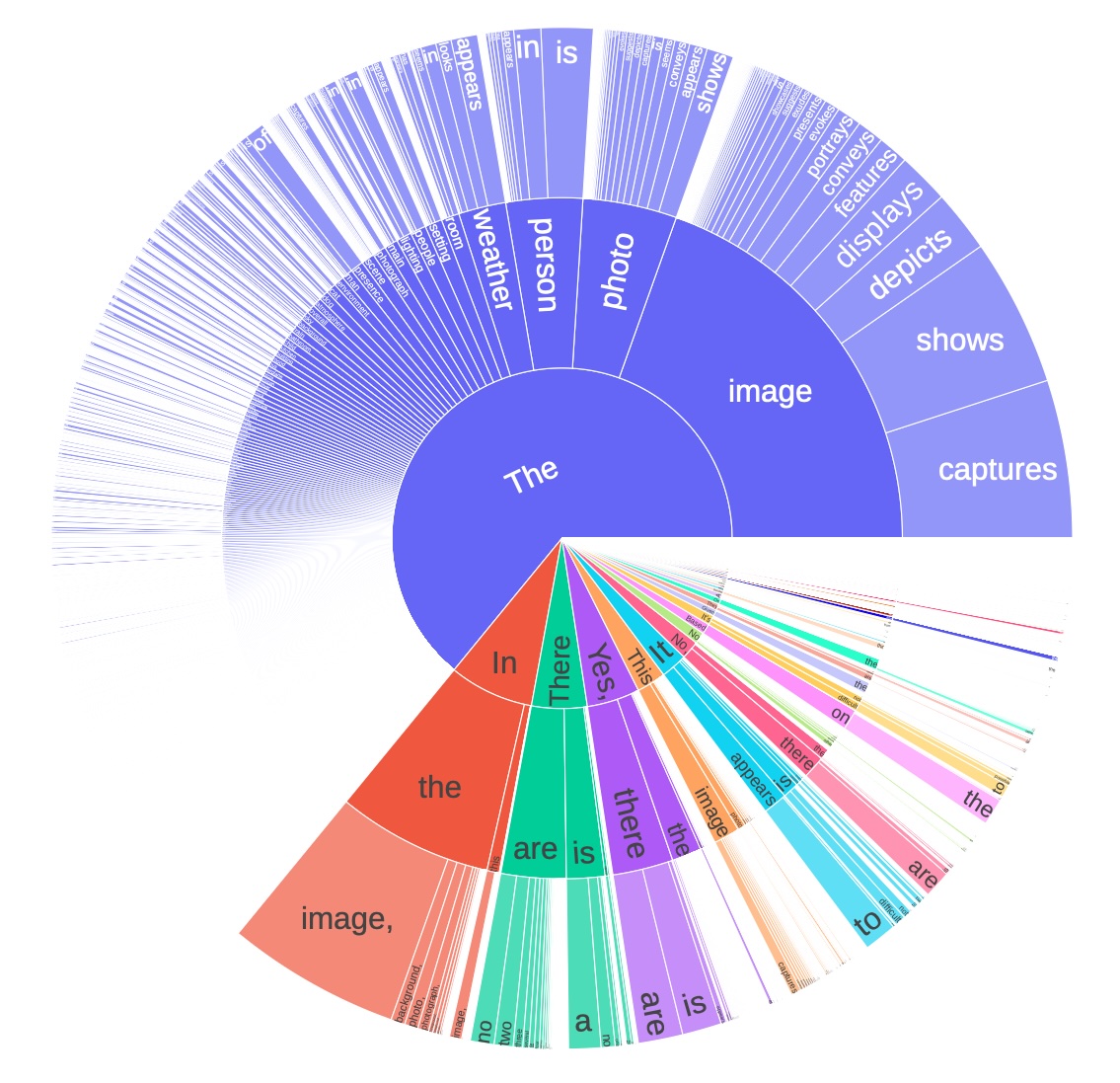}
    \captionsetup{format=plain}
    \caption{
This chart displays the frequency of words found in the \texttt{mvif} answers, organized according to their syntactic order.
  }
    \label{fig:pieplot2}
\end{figure}
\vspace{5cm}
\begin{figure}[h]
    \includegraphics[width=\linewidth]{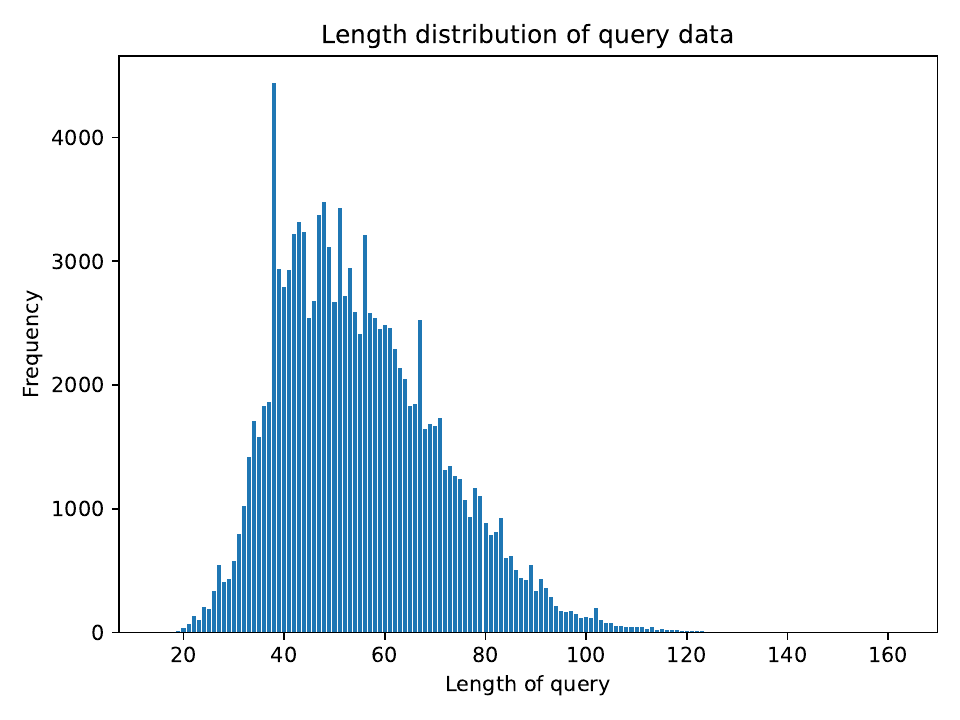}
    \captionsetup{format=plain}
    \caption{
This graph represents the word length distribution of questions in the \texttt{mvif}.
  }
    \label{fig:eq}
\end{figure}

\begin{figure}[h]
    \includegraphics[width=\linewidth]{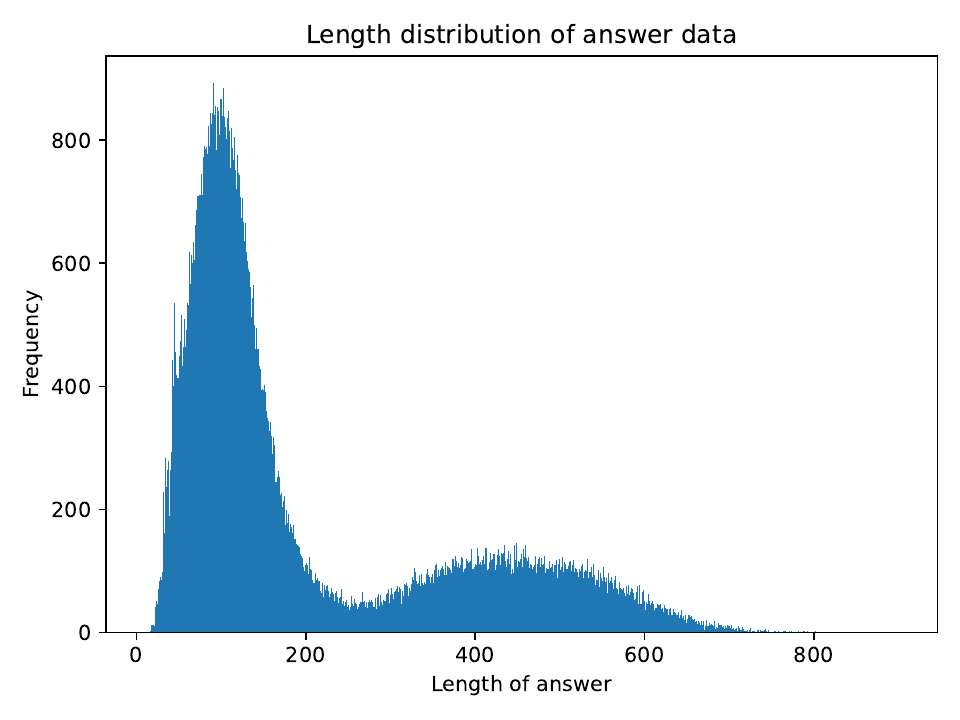}
    \captionsetup{format=plain}
    \caption{
This graph represents the word length distribution of answers in the \texttt{mvif}.
  }
    \label{fig:en_answer_length}
\end{figure}

In this section, we present a detailed analysis of the dataset. Figure~\ref{fig:pieplot1} and Figure~\ref{fig:pieplot2} visualize the frequency distribution of words contained in English questions and answers within the dataset. These graphs follow the order of words in sentences, starting from the center and progressing outward. In other words, the center represents the first word of the sentence, and each subsequent word is represented outwardly based on its position within the sentence.

Figure~\ref{fig:eq} and Figure~\ref{fig:en_answer_length} depict the word lengths of English queries and responses in the context of \texttt{mvif}, providing an overview of the dataset's distribution.

\clearpage
\section{Training Details and Hyperparameters}


\textbf{Training details.} Like LLaVA1.5, we applied Low-Rank Adaptation (LoRA)~\cite{hu2021lora} for visual instruction-following. All the used hyperparameters are identical. Furthermore, we also utilized LoRA in the Korean-English pretraining phase of LLaMA2 to reduce GPU memory usage. The LoRA parameters applied during the pretraining phase were taken directly from the parameter settings suggested in Chinese Alpaca~\cite{cui2023efficient}. \\

\textbf{Training order for VIF data.} We observed significant performance variations depending on the order of the data during the training with VIF data. Therefore, during the visual instruction tuning phase, all data were shuffled and trained together. \\

\textbf{Evaluation Metric for Korean and Chinese.} In this study, the Korean (BVQA, KoLiv, KoViz) and Chinese (VQA-ch) data used differ from English VQA in that they only have one answer per data point. Consequently, answers like ``Yes'', ``네(yes)'', and ``예(yes)'' all have the same meaning, but there is an issue where only ``네'' is counted as the correct answer. Therefore, we conducted post-processing to treat all three responses as correct. We applied the same performance evaluation method across all models. The detailed evaluation script is in our repository: \url{https://github.com/AnonymousMercy/NACCL_submit} \\

\textbf{Hyperparameters.} We employed the same hyperparameter settings as LLaVA1.5. 

\begin{table}[h]
\centering
{\begin{minipage}{0.4\textwidth} 
    \small
    \begin{tabular}{l | c } 
    \hline
{component} & {value} \\ 
\hline
 Dropout & 0.05 \\ 
 Learning rate & 5e-5  \\
 Optimizer & AdamW \\ 
 $\beta_1$, $\beta_2$  & 0.9, 0.99  \\ 
 Epoch for VQA & 1  \\ 
 Batch size (VQA)  & 8  \\
 Low-rank size & 8  \\
 ora\_alpha & 32 \\
 lora\_trainable & {q,v,k,o,gate,down,up}\_proj \\
 LoRA layer,  & q, k, v \\
 {Random Seed} & {42}  \\ 

 \hline
  \end{tabular}
\end{minipage}}
  \caption{Applied hyperparameters.}   \label{tab:hyperparameter}

\end{table}
\newpage
\begin{table}[]
    \centering
    \begin{tabular}{l|llcccc}
    \toprule 
    GPU           & Training Phase     & Duration  \\
    \midrule
    A6000x1    &      CC3M              & 96.6h         \\                              
    A6000x1    &      Wiki-Pretraining(ko) & 28.4h     \\
    A6000x1   &       VIF                  & 64.1h      \\
    \midrule
     Total &                            & 182.1h   \\
    \bottomrule
    \end{tabular}
    \caption{Duration of Each Training Phase. It took a total of approximately 7.5 days to train the proposed enhanced model using the A6000 GPU.}
    \label{tab:gpu}
\end{table}

\section{Generated Data Samples}

\begin{figure}[hbt!]
    \centering
    \includegraphics[width=\linewidth]{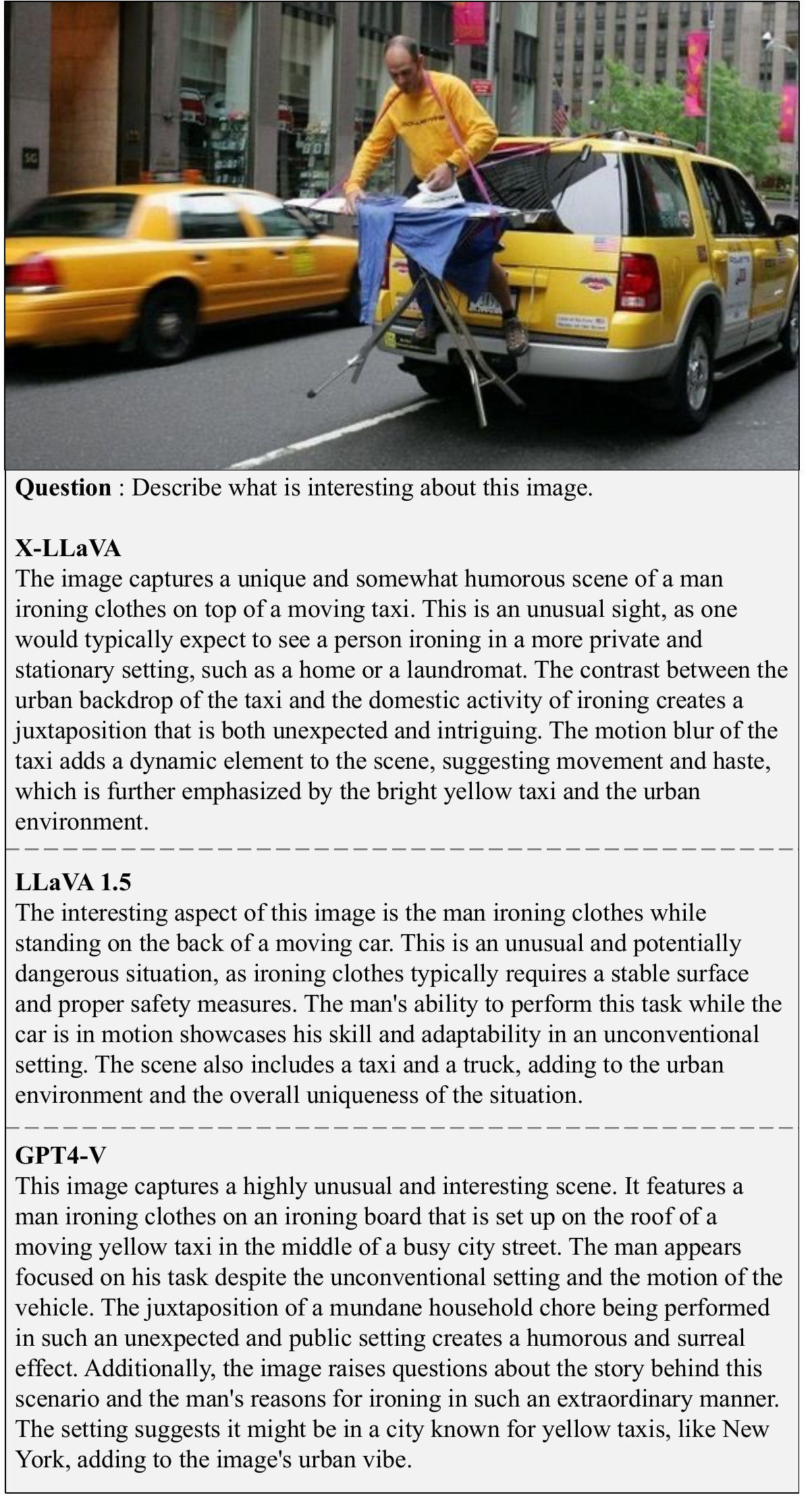}
    \caption{Example of each model's answer to ``Describe what is interesting about the image''}
    \label{fig:enter-label}
\end{figure}
\clearpage

\begin{figure*}
    \centering
    \includegraphics[height=1.5\linewidth]{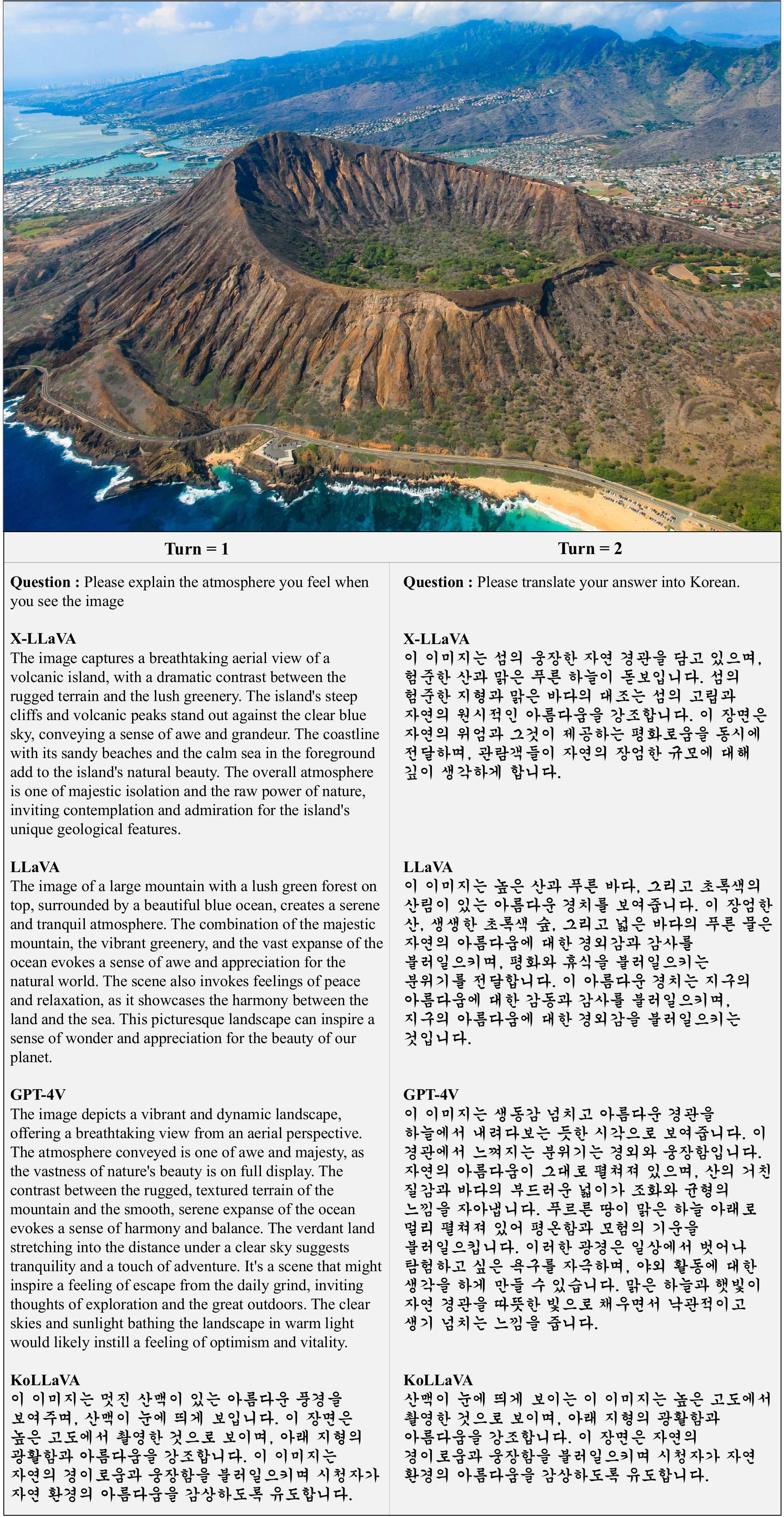}
    \caption{The results of multi-turn conversations across various models.}
    \label{fig:enter-label}
\end{figure*}

\clearpage

\section{Statistical Analysis of Responses in the Qualitative Evaluation Experiment}
\begin{figure}[hbt!]
    \centering
    \includegraphics[width=\linewidth]{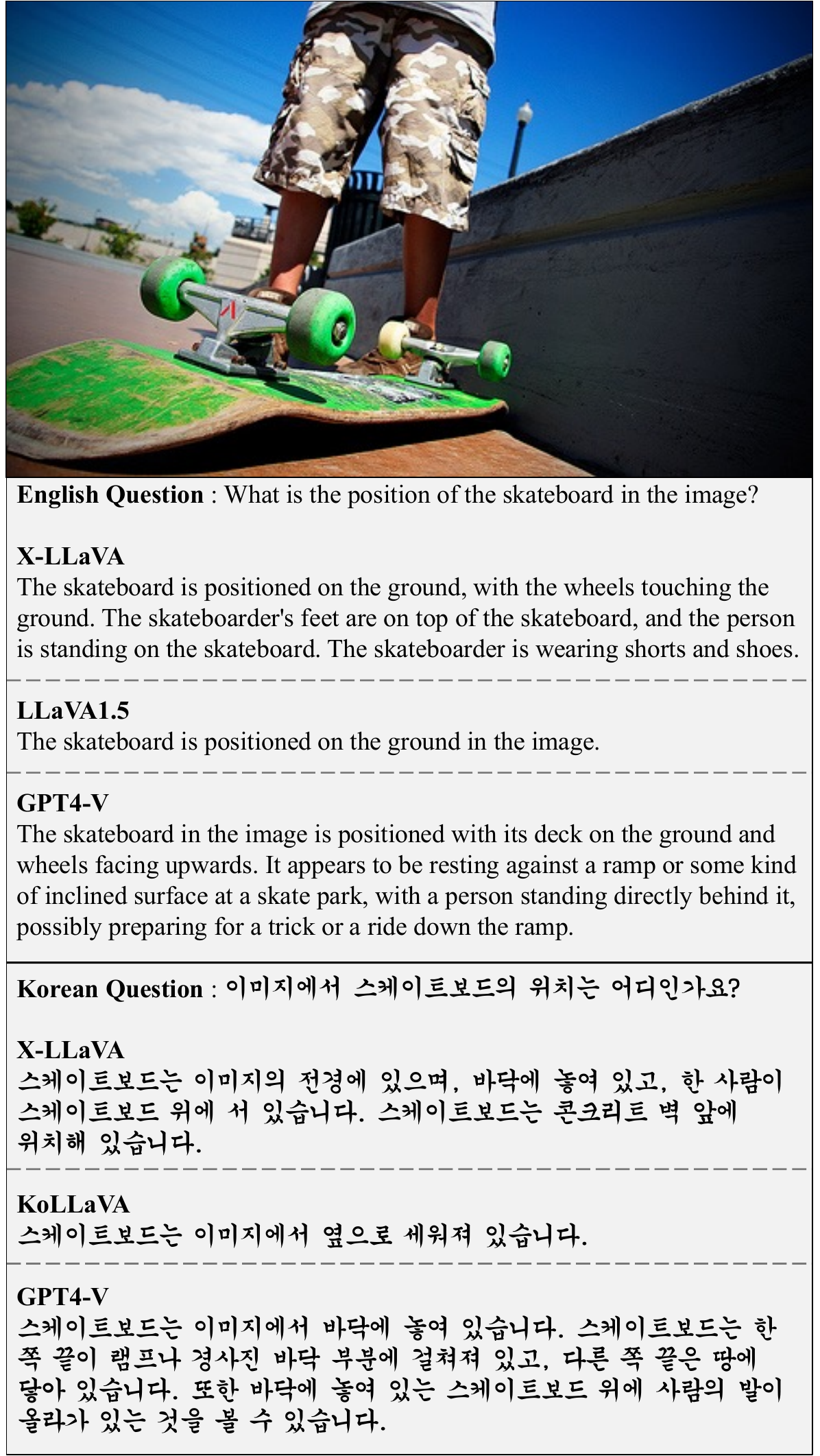}
    \scriptsize\caption{Examples of answers from `X-LLaVA', `LLaVA', `KoLLaVA', and `GPT4-V' models to English and Korean qualitative evaluations.}
    \label{fig:qu_eval_exmaple1}
\end{figure}

\begin{figure}[hbt!]
  \includegraphics[width=\linewidth]{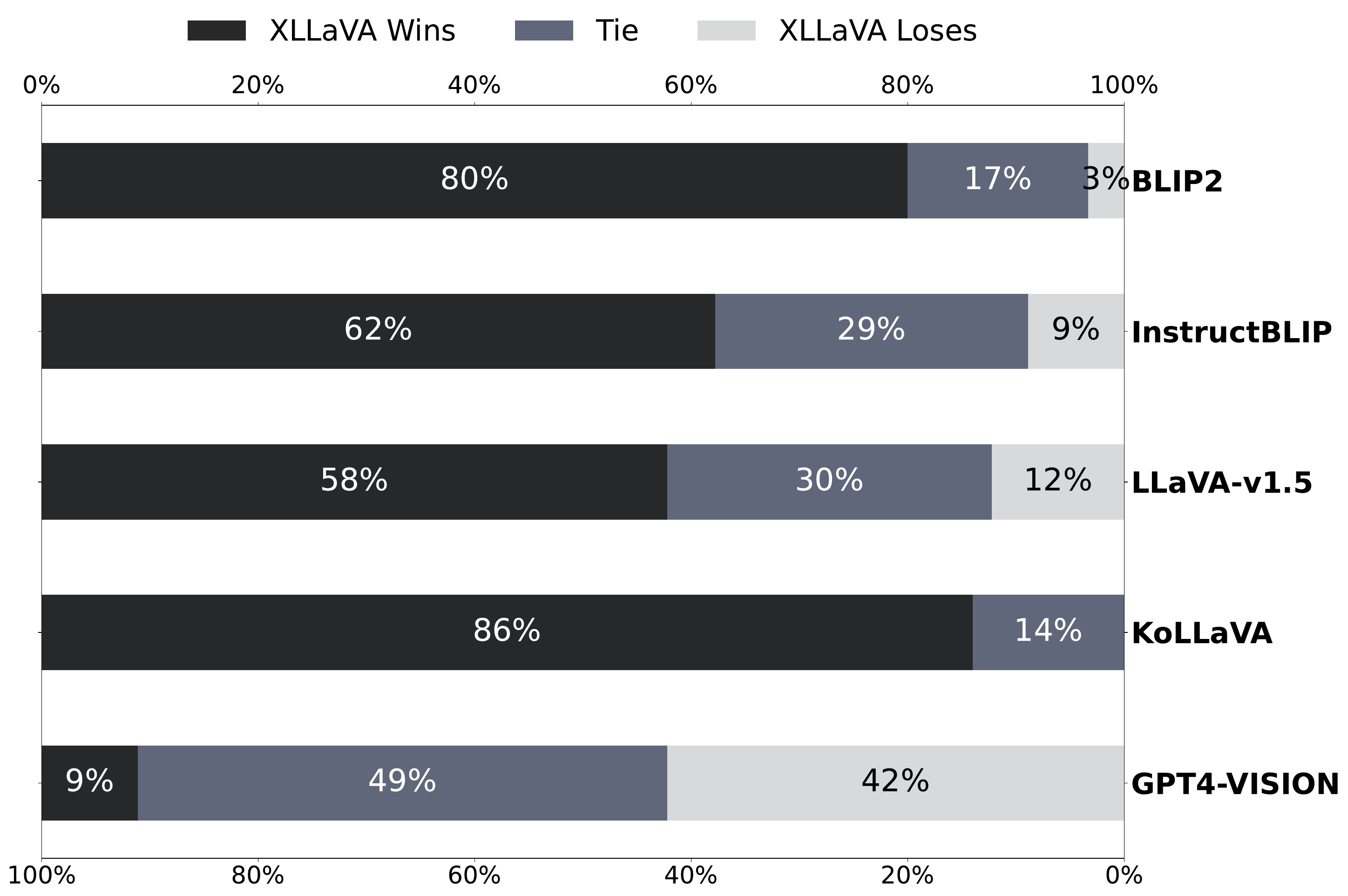}
  \captionsetup{format=plain}
  \caption{Results of GPT preference evaluation for English when limited to 30 words.}
  \label{fig:qualitative Evaluation (EN, short)}
\end{figure}

\begin{table}[hbt!]
    \centering
    \small
    \begin{tabular}{lcccccc}
    \toprule
    POS & \multicolumn{2}{c}{X-LLaVA}  & \multicolumn{2}{c}{KoLLaVA} & \multicolumn{2}{c}{GPT4-V} \\
    \midrule
    Duplicate & \cmark & \xmark & \cmark & \xmark & \cmark & \xmark\\
    \cmidrule(lr){2-3} \cmidrule(lr){4-5} \cmidrule(lr){6-7}
    Noun & 899 & 414 & 266 & 133 & 879 & 455\\
    Verb & 247 & 59 & 39 & 19 & 251 & 74 \\
    Modifier & 67 & 14 & 30 & 10 & 64 & 19 \\
    Indep. & 2 & 2 & 10 & 2 & 4 & 2 \\
    Relational & 533 & 23 & 89 & 15 & 498 & 27 \\
    Ending & 366 & 21 & 51 & 16 & 370 & 29 \\
    Affix & 49 & 4 & 6 & 4 & 61 & 7 \\
    Symbols & 138 & 3 & 106 & 5 & 142 & 2 \\
    F.L. & 2 & 1 & 35 & 1 & 0 & 0 \\
    Total & 2303 & 541 & 632 & 205 & 2269 & 615 \\
    \bottomrule
    \end{tabular}
    \caption{It is to compare the number of parts of speech based on the Korean answers to the qualitative evaluation that limited 30 words, and `Duplicate' means whether or not words are duplicated. In the following table, `Part Of Speech' is specified as `POS', `Independent' is specified as `Indep.', and `Foreign Language' is specified as `F.L.'}
    \label{tab:my_label}
\end{table}

\textbf{Analysis of Qualitative Evaluation} Figure~\ref{fig:qu_eval_exmaple1} shows examples of responses from each model in the qualitative evaluation. When examining the responses of X-LLaVA, it is noticeable that there is a tendency to focus on the positions of objects. This indicates that X-LLaVA has been trained on the \texttt{mvif} dataset, which includes a variety of tasks involving Objects and Locations. Additionally, as observed in Figure~\ref{fig:qualitative Evaluation (EN, short)}, X-LLaVA outperformed all other models except for the GPT4-V model in the evaluations. Particularly, as shown in Table~\ref{tab:my_label}, X-LLaVA used a more diverse and extensive range of expressions compared to KoLLaVA, which likely had a significant impact on the comparisons in Figure~\ref{fig:qualitative Evaluation (EN, short)}. This suggests that the LLM vocab expansion technique employed in training X-LLaVA contributed to its effectiveness. However, despite using a variety of expressions similar to GPT4-V, X-LLaVA was outperformed by GPT4-V by a margin of 33\%, implying that GPT4-V likely used more implicit and advanced vocabulary within shorter sentences.

\begin{figure}[ht!]
    \includegraphics[width=\columnwidth]{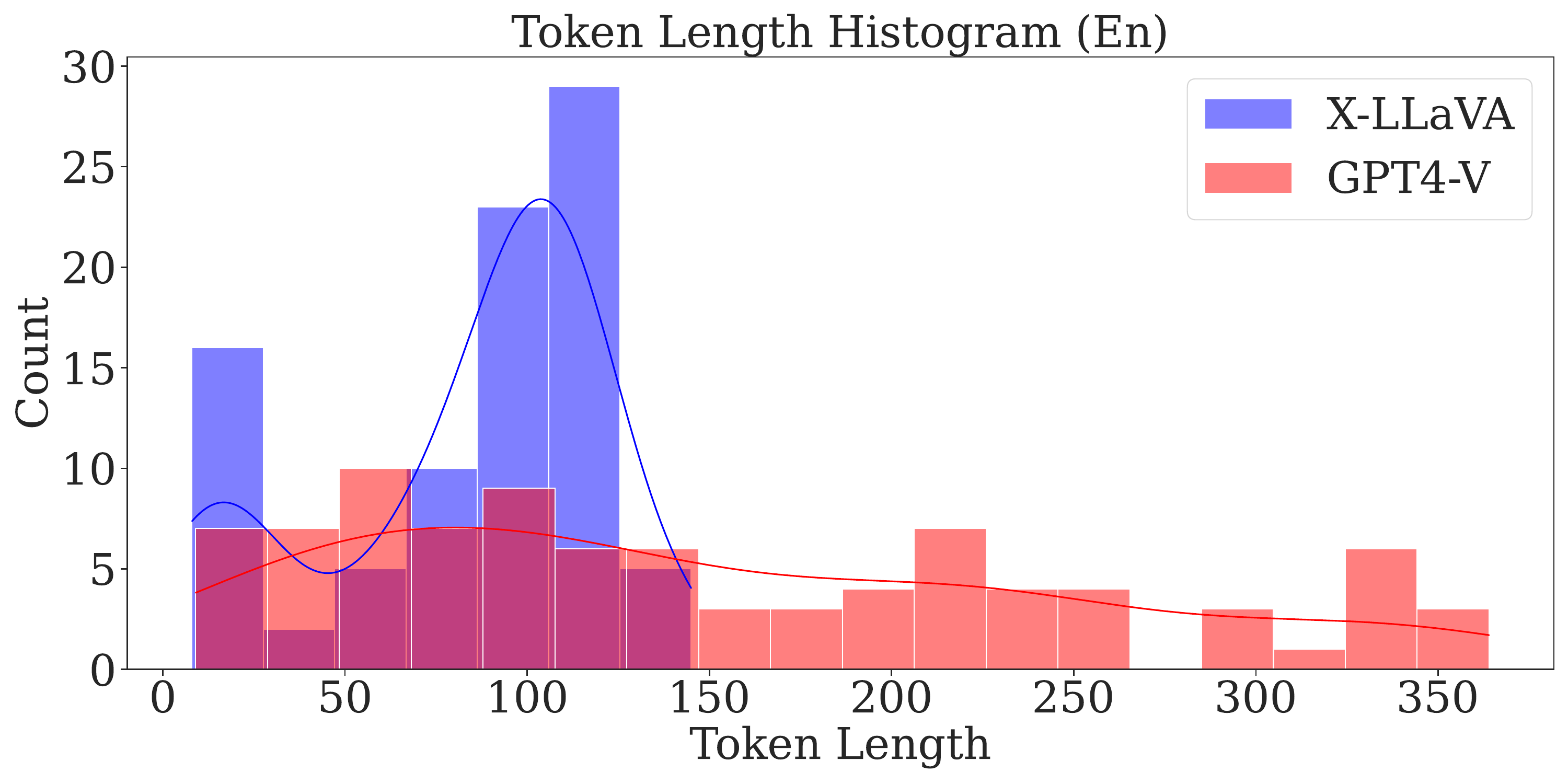}
    \captionsetup{skip=0.3cm}
    \scriptsize\caption{
   This graph shows a histogram comparing the token lengths of English answers for `X-LLaVA' and `GPT4-V'. }
   \label{fig:our_gpt_en}
\end{figure}
\vspace{-0.4cm}

\begin{figure}[ht!]
    \includegraphics[width=\columnwidth]{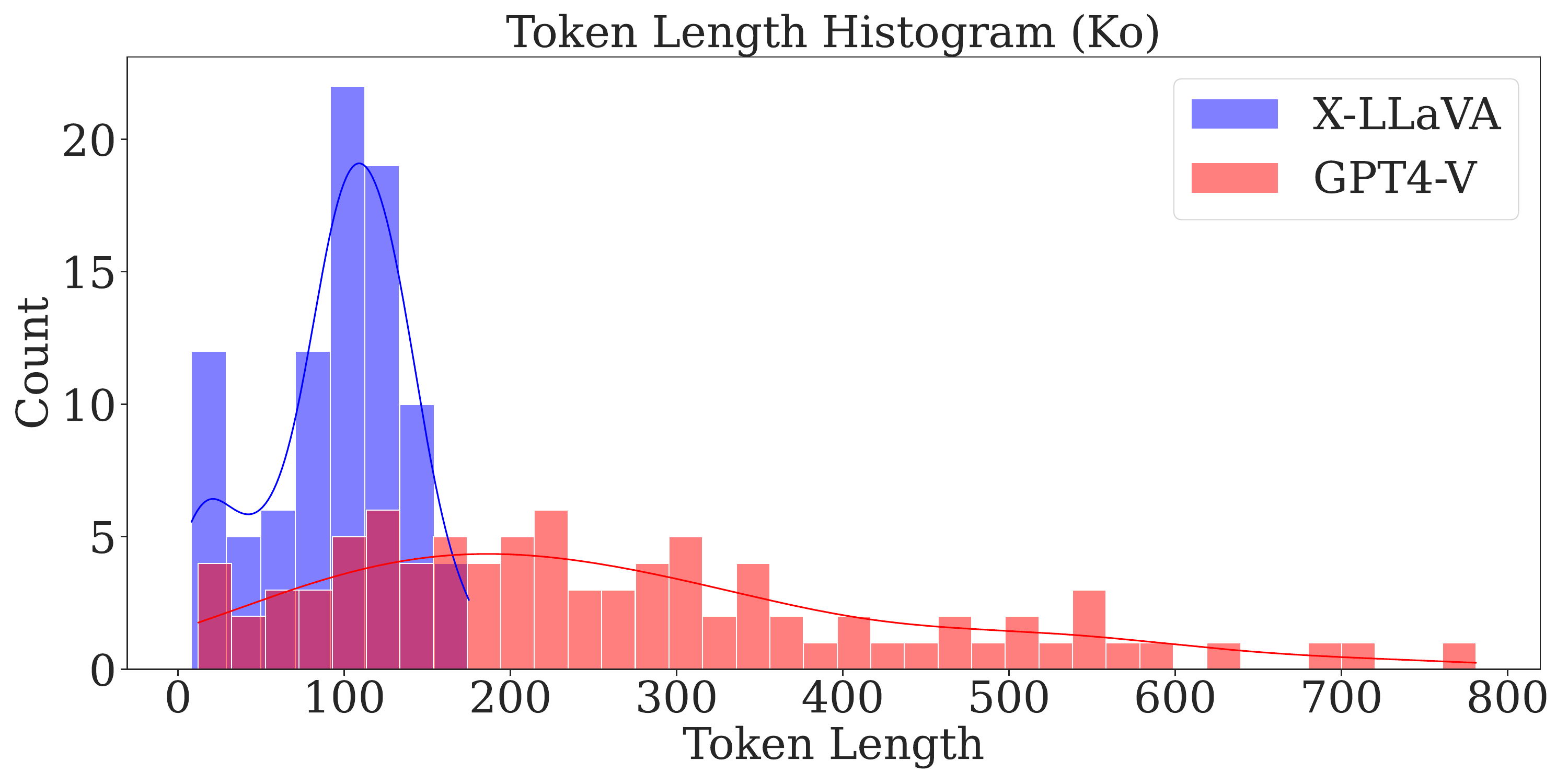}
    \scriptsize
    \captionsetup{skip=0.3cm}
    \caption{
   This graph shows a histogram comparing the token lengths of Korean answers for `X-LLaVA' and `GPT4-V'.}
   \label{fig:out_gpt_ko}
\end{figure}
\vspace{-0.4cm}

\begin{figure}[ht!]
    \includegraphics[width=\columnwidth]{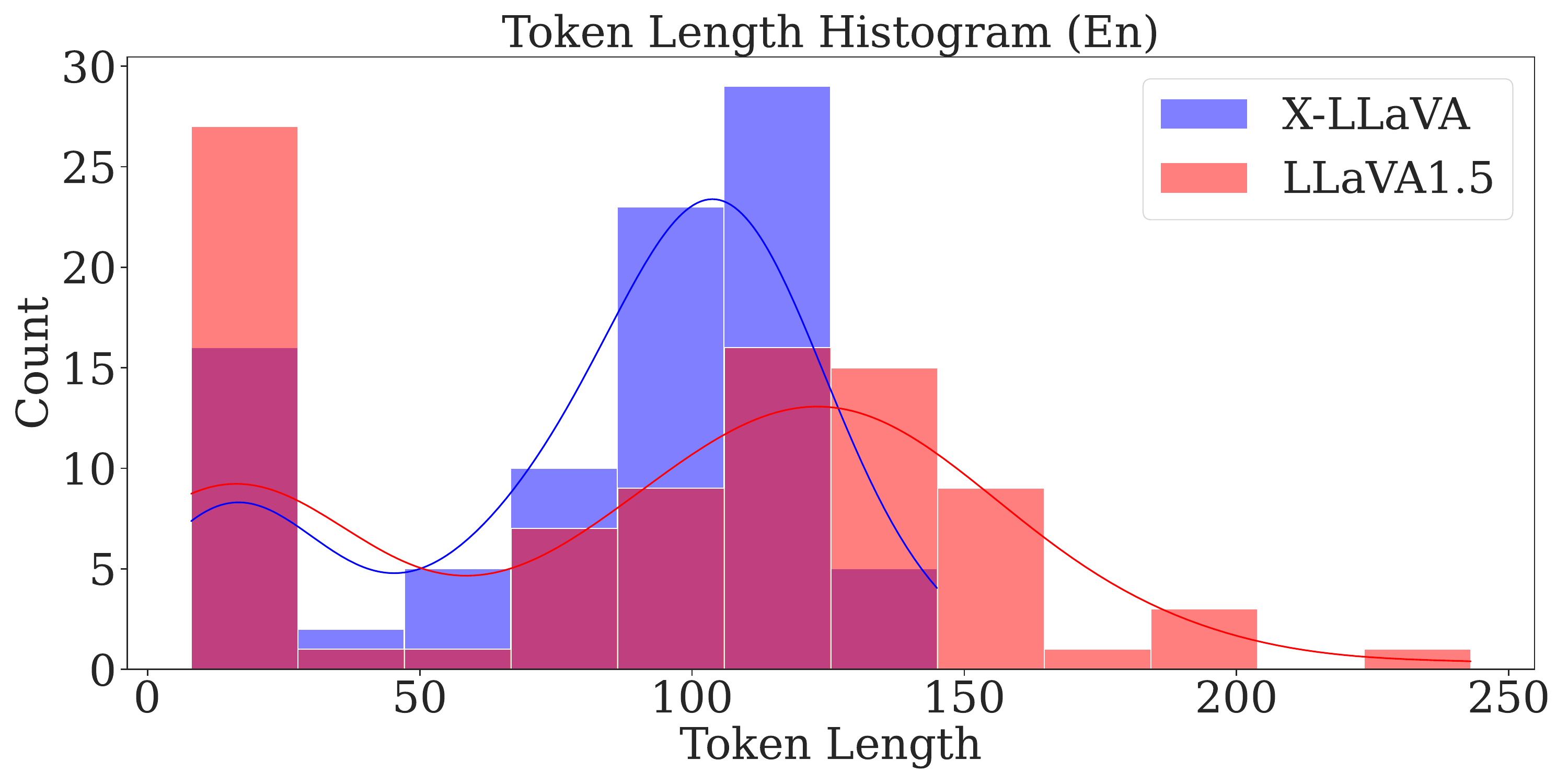}
    \scriptsize
    \captionsetup{skip=0.3cm}
    
    \caption{
    This graph shows a histogram comparing the token lengths of English answers for `X-LLaVA' and `LLaVA1.5'.}
    \label{fig:our_llava15}
\end{figure}
\vspace{-0.4cm}

\begin{figure}[ht!]
    \includegraphics[width=\linewidth]{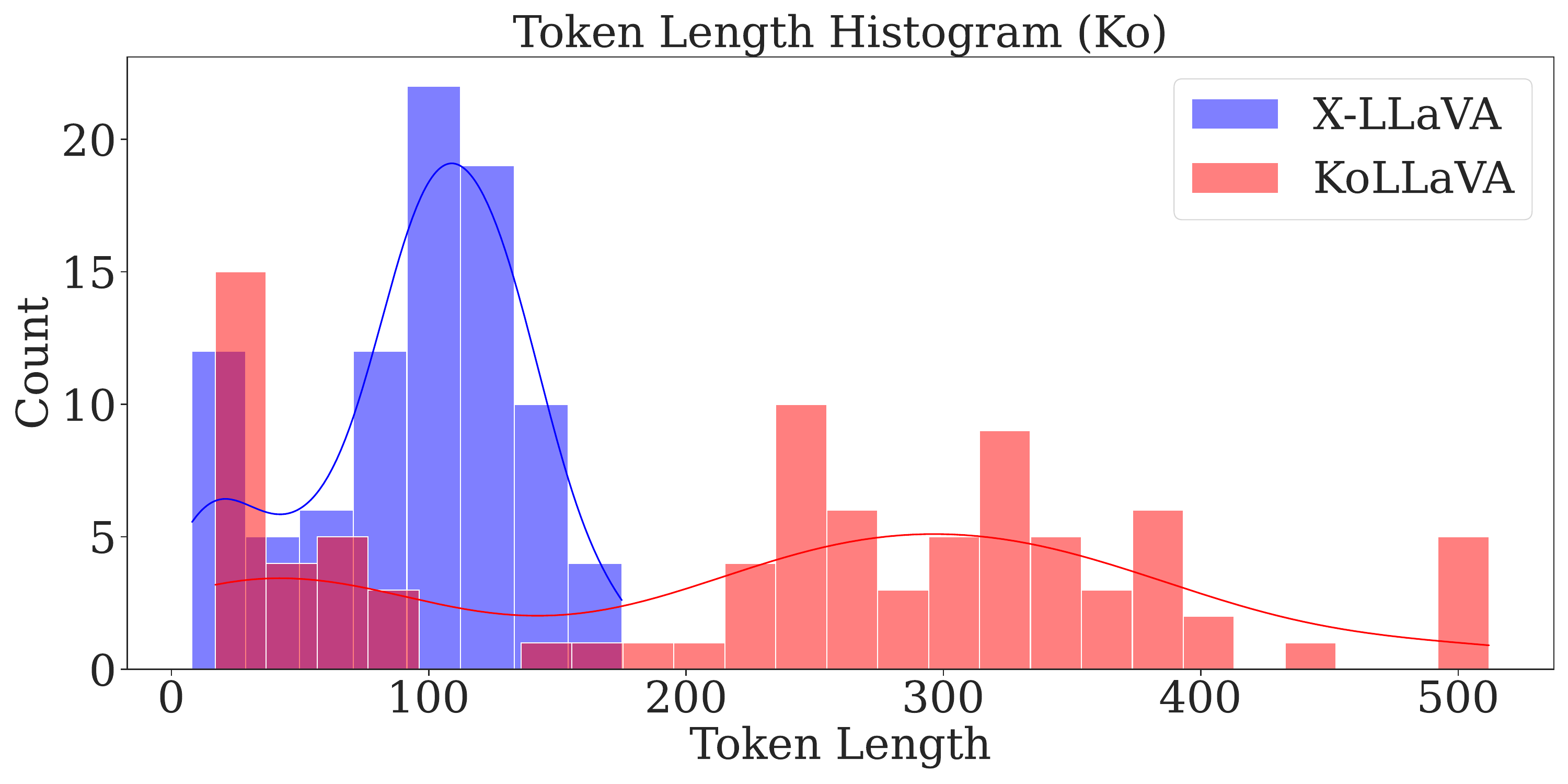}
    \scriptsize
    \captionsetup{skip=0.3cm}
    \caption{
    This graph shows a histogram comparing the token lengths of Korean answers for `X-LLaVA' and `KoLLaVA'.}
    \label{fig:our_kollava}
\end{figure}

\newpage
Figure \ref{fig:our_gpt_en}$\sim$\ref{fig:our_kollava} visualize the distribution of token lengths In this study, the model proposed, X-LLaVA, tends to produce relatively shorter responses. When contrasted with the results of the GPT4-V qualitative evaluation, Figure~\ref{fig:our_llava15} shows that LLaVA1.5, while having a distribution of response lengths similar to other models, has a lower win rate than X-LLaVA, suggesting that the X-LLaVA model generally produces higher quality English responses than the LLaVA1.5 model. Additionally, Figure~\ref{fig:our_kollava} shows that KoLLaVA, despite generally having longer responses than X-LLaVA, has a relatively lower win rate. This indicates a tendency of the X-LLaVA model to generate higher quality Korean responses relative to the same response length.

\section{Human Preferenece Evaluation Details}

\begin{table}[]
    \small
    \centering
    \begin{tabular}{l|cccc}
    \toprule 
     Evaluator & XLLaVA Wins & Tie & XLLaVA Loses \\
    \midrule 
    GPT4-V(G) & 15 & 37 & 38      \\            
    Human(H) & 37 & 14 & 39 \\
    \midrule
    G~$\cap$~H & 12 & 10 & 32 \\ 
    \bottomrule
    \end{tabular}
    \caption{It displays the number of samples chosen by GPT4-V and Human Evaluators for `XLLaVA Wins', `Tie', and `XLLaVA Loses', respectively in Figure~\ref{fig:qualitative Evaluation (KO, short)} and~\ref{fig:kor_llm_eval_human}. `G~$\cap$~H' signifies instances where both evaluators (Human, GPT4-V) indicate the same outcome for each of the 90 samples.}
    \label{tab:gpt4_v_equal_human}
\end{table}

\begin{figure}[]
\centering
\includegraphics[width=\linewidth]{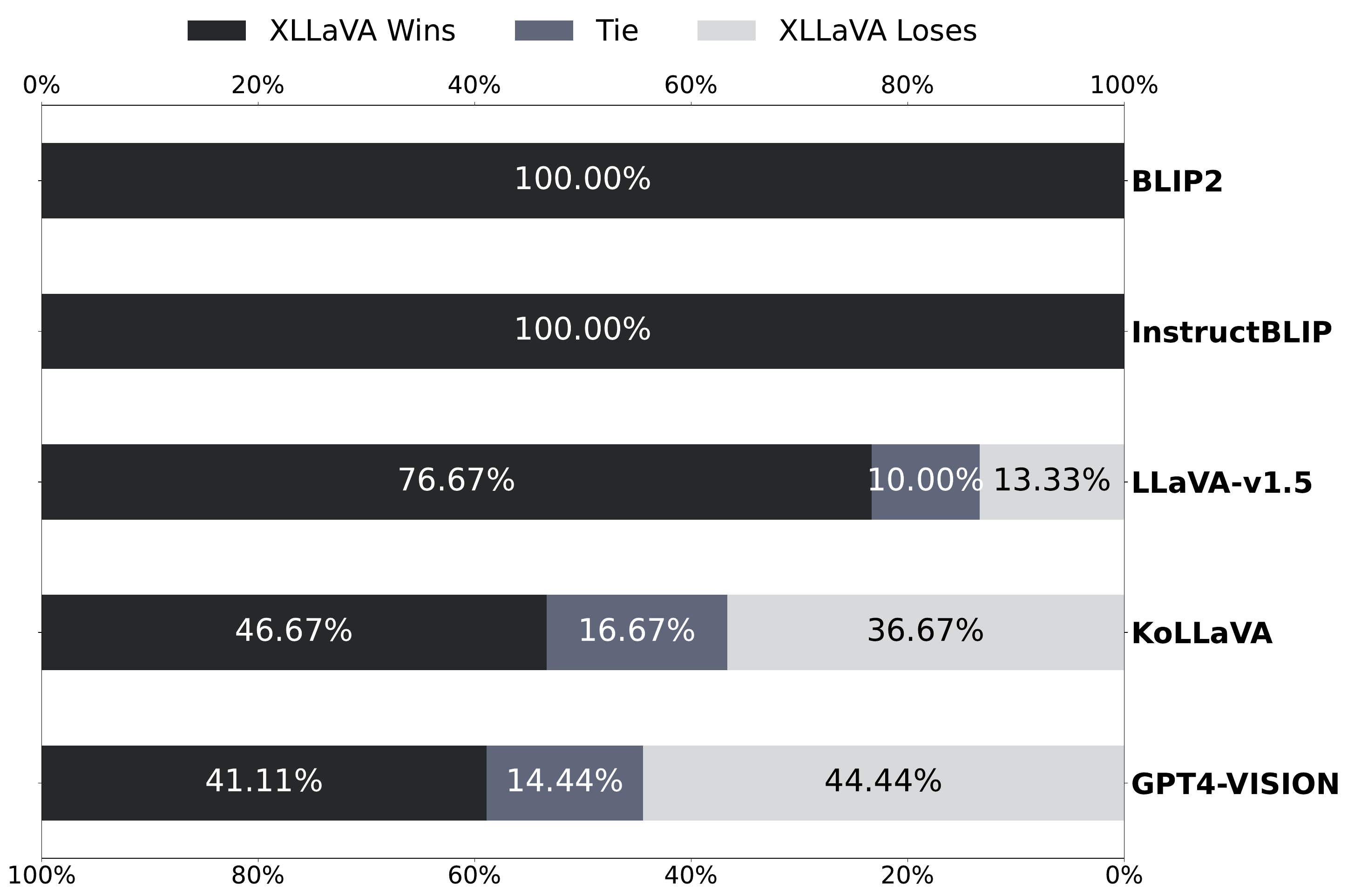}
\caption{Preference evaluation by human in comparison with other models}
\label{fig:kor_llm_eval_human_all}
\end{figure}

The Human Preference Evaluation shown in Figure~\ref{fig:kor_llm_eval_human} was carried out with three evaluators using the following criteria: For a result to be classified as `XLLaVA Wins,' either all three evaluators needed to select it or at least two did. A `Tie' was determined either when all evaluators agreed on it or when their selections were evenly split across `XLLaVA Wins,' `Tie,' and `XLLaVA Loses.' Similarly, `XLLaVA Loses' was classified when all three agreed on it or at least two of the three chose it.
Table~\ref{tab:gpt4_v_equal_human} presents the numerical results corresponding to those depicted in Figure 6. The evaluation results between Human and GPT4-V show an 80\%(12/15) agreement rate for `XLLaVA Wins' and approximately an 82\%(32/38) agreement rate for `XLLaVA Loses (GPT4-V Wins)'. However, for the `Tie' category, the GPT4-V Evaluation only shows about a 27\% agreement rate with human evaluations, indicating a significant difference compared to the results of Human Preference Evaluation. Therefore, at this stage, it is challenging for GPT preference evaluations to serve as a complete substitute for human assessments. Nonetheless, the overarching trends observed in these preference evaluations bear some resemblance to those in human assessments, suggesting that they constitute a meaningful metric for consideration.

We also extended our evaluations to include models other than X-LLaVA, employing the same human evaluation protocol. Figure~\ref{fig:kor_llm_eval_human_all} displays the human evaluation results in Korean for all models examined in this study. Consistent with previous discussions, while the overall trend in GPT and human evaluations across different models was generally similar, GPT was more prone to result in ties in preference assessments.

A notable aspect of this experiment is that, in contrast to the GPT evaluations, X-LLaVA achieved a complete win in all trials against BLIP2 and InstructBLIP2, models that lack proficiency in Korean. Conversely, the GPT evaluations depicted in Figure~\ref{fig:qualitative Evaluation (KO, short)} resulted in a ``Tie'' for 7 to 17\% of the cases involving these two models, which also do not understand Korean. This pattern indicates that GPT adopts a highly conservative approach in its evaluations, potentially due to its methodology or criteria for determining outcomes, emphasizing caution and possibly erring towards neutrality when faced with ambiguous cases.

\section{Inspection Procedure Details}

\begin{figure}[hbt!]
    \centering
\includegraphics[width=0.8\linewidth]{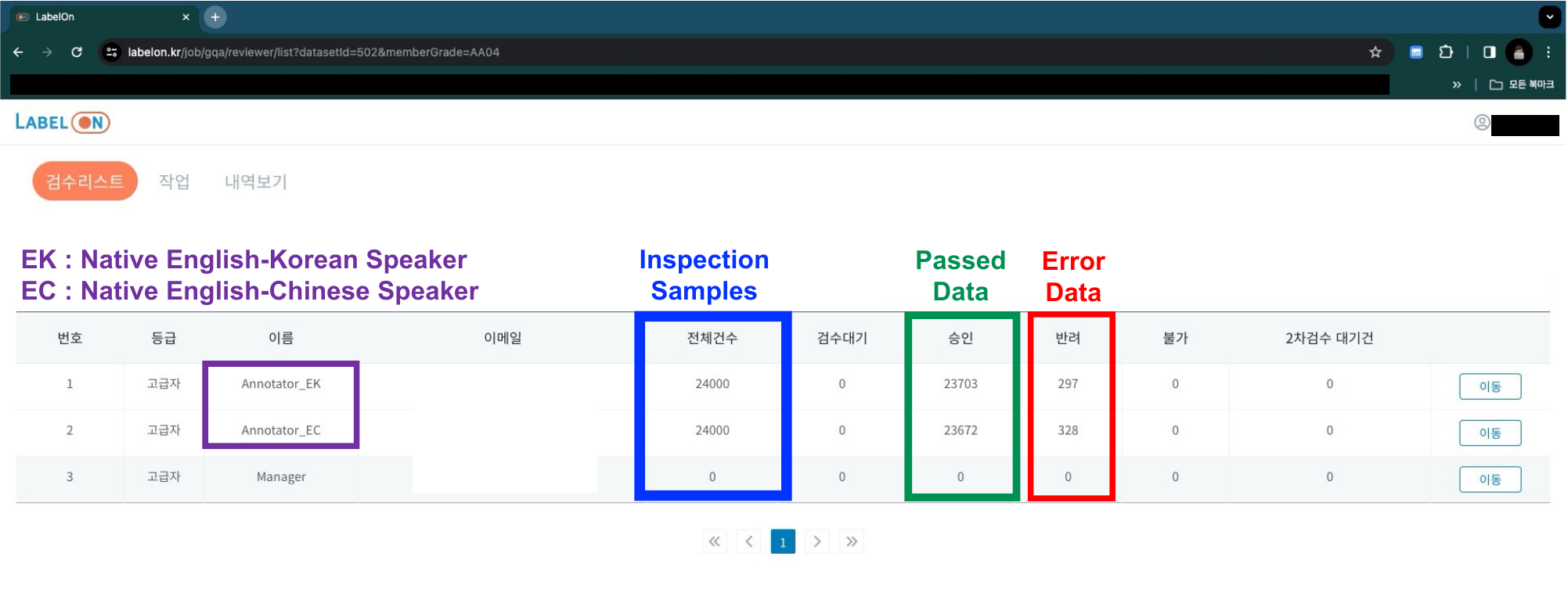}
    \scriptsize\caption{
    This figure represents the worker status board on the LabelON data review platform. Information about the annotators is shown in purple, the work target samples in blue, Passed Data in green, and Error Data in red.
  }
    \label{fig:labelon1}
\end{figure}

\begin{figure}[ht!]
    \centering

    \begin{subfigure}[b]{\linewidth}
        \centering
        \includegraphics[width=0.8\linewidth]{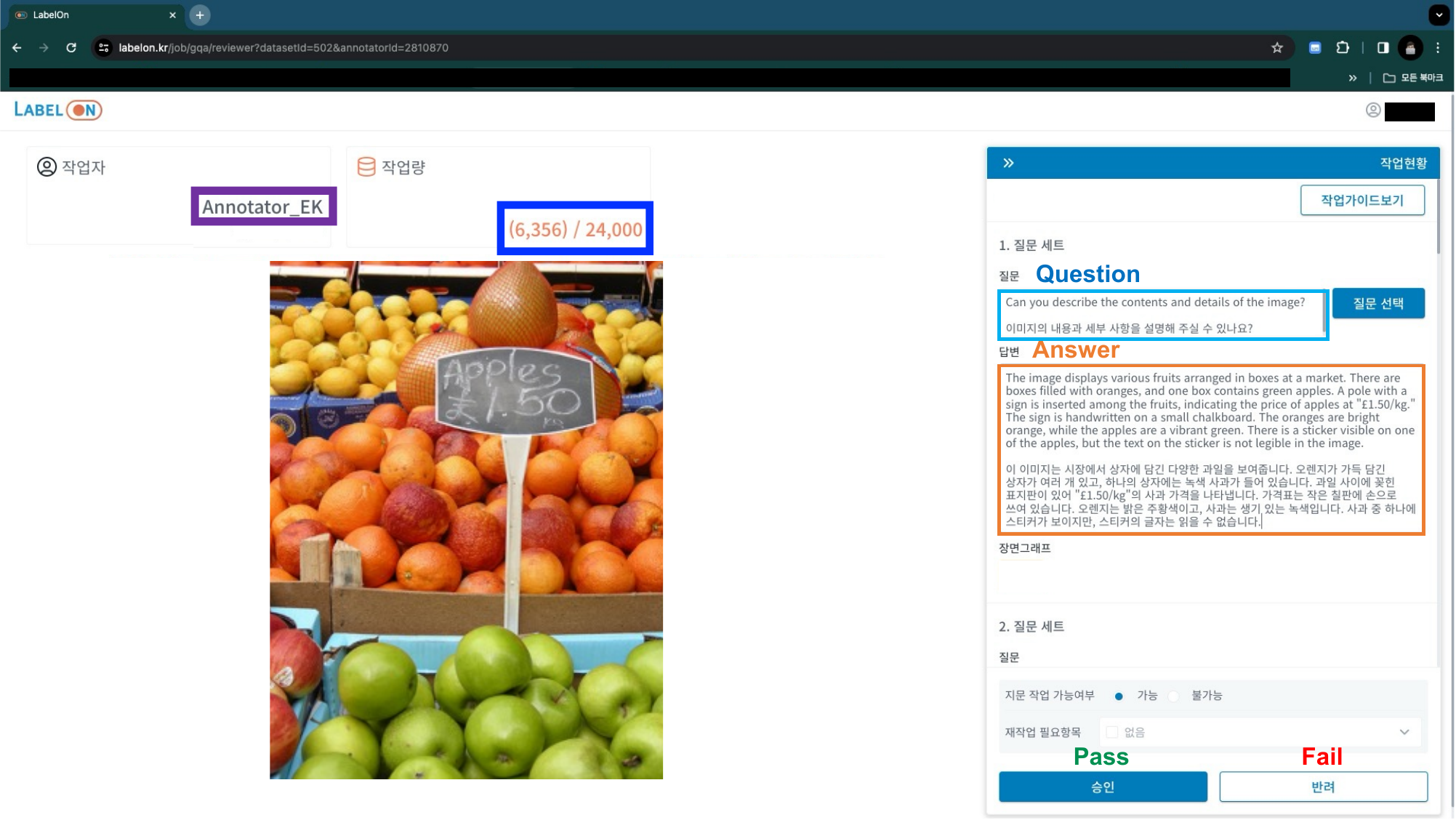}
        \caption{English-Chinese dataset inspection procedure}
        \label{fig:labelon2}
    \end{subfigure}

    \vspace{1em} 

    \begin{subfigure}[b]{\linewidth}
        \centering
        \includegraphics[width=0.8\linewidth]{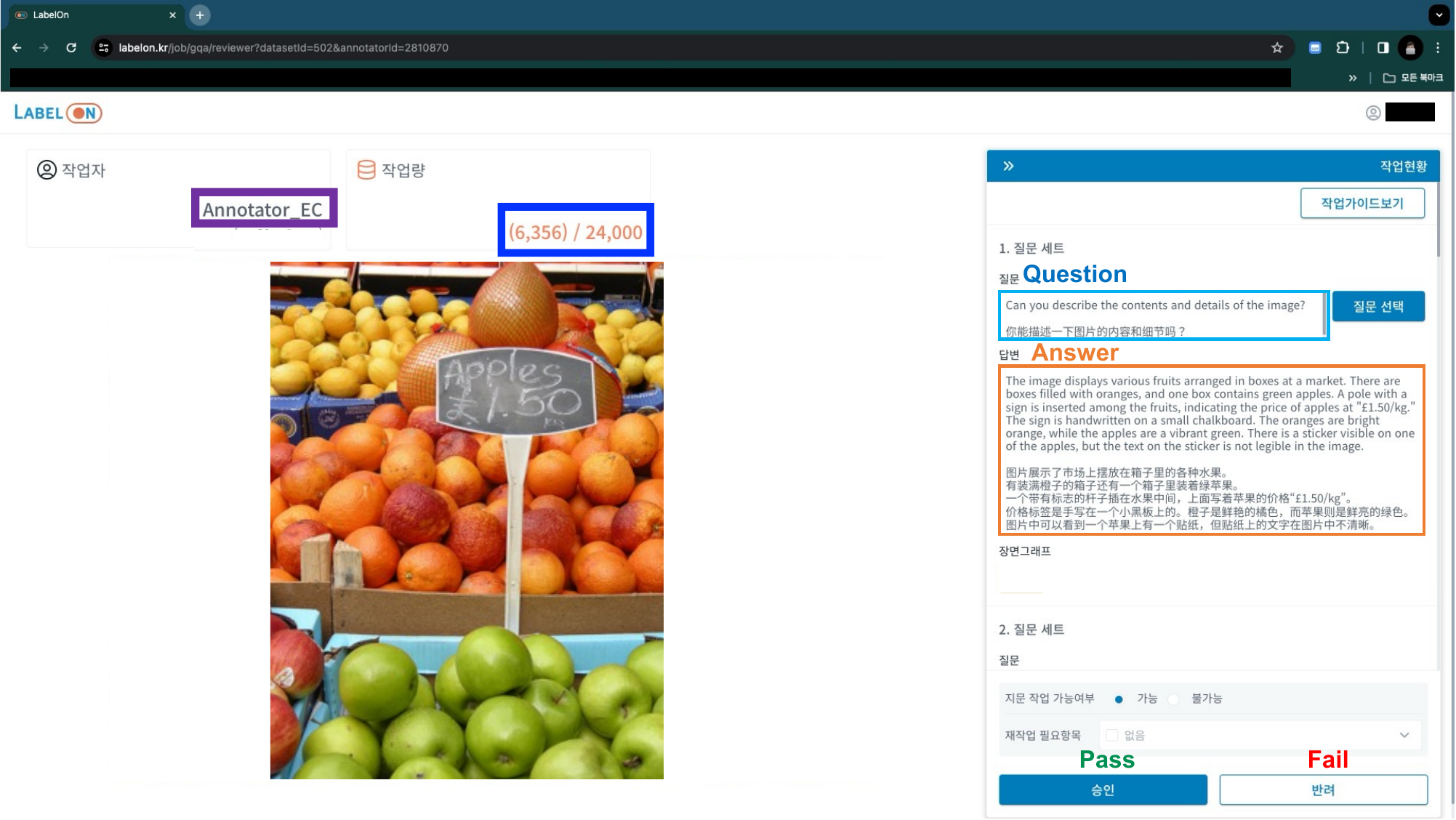}
        \caption{English-Korean dataset inspection procedure}
        \label{fig:labelon3}
    \end{subfigure}
    
    \caption{It shows the workflow on the LabelON data review platform. Information about annotators is displayed in purple, Questions in sky blue, Answers in orange, Passed annotations in green, and Errors in red.}
    \label{fig:label_on}
\end{figure}

We have employed two annotators, one native English-Korean speaker and one native English-Chinese speaker, to inspect the generated data for 24,000 images. To facilitate efficient data inspection, we utilized a WebUI-based data inspection platform (LabelOn), where annotations can be verified through Figure~\ref{fig:labelon1} and ~\ref{fig:label_on}. Each annotator received parallel sets of English-Chinese or English-Korean datasets to review for Pass/Error statuses. Both annotators inspected the data over the course of one month. As a result, 504 data points were removed. The two main issues with the removed data were identified as (1) proper noun objects and (2) cultural differences. Below are examples:\\

\textbf{(1) For the issue of proper noun objects:}
\begin{itemize}
    \item Question: Describe the scene in the image
    \item Answer: “..north ridge of Mount Stuart..”
\end{itemize}

In cases like the above, GPT4-V labeled the location with the proper noun “Mount Stuart” based on its own knowledge despite it being difficult to specify the place from the input image. Such data were problematic and, therefore, deleted.\\

\textbf{(2) For the issue of cultural differences:} 
We found that GPT4-V is also biased towards English-speaking cultures. For example,

\begin{itemize}
    \item Question: Describe the scene in the image
    \item Answer: “…. creepy food ….”
\end{itemize}
`Creepy food' is usually associated with Halloween foods and was translated into Korean and Chinese as “소름끼치는 음식 (scared food)” and “惊悚食物 (thriller food)”, respectively. This is not only a rarely used expression in Korea and China but also has the potential for mistranslation. In this paper, we removed the 504 training data with the issues mentioned above and shared both the original 24K dataset and the post-processed (final) dataset of 23.4K. 

\newpage
\onecolumn

\end{document}